\documentclass{svproc}
\usepackage{url}

\usepackage{amsmath,amsfonts,amssymb,amsthm,epsfig,epstopdf,array}
\newtheorem{assumption}{Assumption}
\usepackage[english]{babel}
\usepackage[utf8]{inputenc}
\usepackage{amsmath}
\usepackage{float}
\usepackage{graphicx,subfigure}
\usepackage[utf8]{inputenc}
\usepackage[english]{babel}
\usepackage{amsfonts}
\usepackage{comment} 
\usepackage{tikz}
\usepackage{mathtools}
\usepackage{fixltx2e}
\usepackage{pifont}
\usepackage{caption}
\usepackage[colorinlistoftodos]{todonotes}
\usepackage{amsmath}
\usepackage{amsfonts}
\usepackage{mathtools}
\usepackage{graphicx}
\usepackage{algpseudocode} 
\usepackage{algorithm}
\usepackage{amssymb,amsmath,amsthm, amsfonts}
\newtheorem{mydef}{Problem}
\usepackage{amssymb,amsmath,amsthm}
\usepackage[linkcolor=blue,colorlinks=true]{hyperref}
\newcommand{\norm}[1]{\left\lVert#1\right\rVert}

\begin{document}
\mainmatter              
\title{Distributed Multirobot Control for Non-Cooperative Herding}
\titlerunning{Distributed Herding}  %
\author{Nishant Mohanty$^{*}$, Jaskaran Grover$^{*}$, Changliu Liu, Katia Sycara}
\institute{The Robotics Institute, Carnegie Mellon University, Pittsburgh, USA\\
\email{\{nishantm,jaskarag,cliu6,sycara\}@andrew.cmu.edu}
\def\thefootnote{$^{*}$}\footnotetext{These authors contributed equally to this work.}\def\thefootnote{\arabic{footnote}}
}
\maketitle              
\begin{abstract}
In this paper, we consider the problem of protecting a high-value area from being breached by \textit{sheep} agents by crafting motions for \textit{dog} robots. We use control barrier functions to pose constraints on the dogs' velocities that induce repulsions in the sheep relative to the high-value area. This paper extends the results developed in our prior work on the same topic in three ways. Firstly, we implement and validate our previously developed centralized herding algorithm on many robots. We show herding of up to five sheep agents using three dog robots. Secondly, as an extension to the centralized approach, we develop two distributed herding algorithms, one favoring feasibility while the other favoring optimality. In the first algorithm, we allocate a unique sheep to a unique dog, making that dog responsible for herding its allocated sheep away from the protected zone. We provide feasibility proof for this approach, along with numerical simulations. In the second algorithm, we develop an iterative distributed reformulation of the centralized algorithm, which inherits the optimality (\textit{i.e.} budget efficiency) from the centralized approach. Lastly, we conduct real-world experiments of these distributed algorithms and demonstrate herding of up to five sheep agents using five dog robots. Videos of these results are available at \url{https://bit.ly/3bZq0dB}.
\end{abstract}

\section{Introduction}
\label{sec:1}

\let\thefootnote\relax\footnotetext{This research is supported by AFOSR FA9550-18-1-0097 and AFRL/AFOSR
FA9550-18-1-0251}

Recent developments in robotics and sensing have created significant interest among researchers to deploy multiple robots to operate cooperatively towards achieving a common goal. Many works have developed techniques to tackle real-world problems using multi-robot systems (MRS), like conducting surveys or automating warehouses  \cite{d2012guest,d2003distributed}, \cite{kazmi2011adaptive}. The major developments in MRS for enabling multiple robots to behave cooperatively have been based on interactions within a single team, i.e., a robot interacts with other robots in its group to achieve a given objective \cite{ji2007distributed,lin2004multi}. The main features of these types of algorithms are a) local interaction, b) collision-free motion within the group, and c) achieving collective behavior using local interaction \cite{reynolds1987flocks}.  

In literature, there are studies on MRS that involve interaction between multiple groups of agents. Here, along with the local interaction with group members, the individuals also interact with an external agent from another group. An example of this is a scenario where a group of adversarial robots has a goal of their own that might damage a given high-value unit. Here, a group of defenders must interact with the adversarial robots to ensure the safety of the unit. \cite{walton2021defense,tsatsanifos2021modeling}.
In this paper, we propose a provably correct controller for the group of defenders  (``dog robots") to prevent an adversarial group  (the ``sheep robots") from breaching a protected zone. This is challenging because dog robots do not control the sheep robots directly; rather have to rely on the interaction dynamics between the dogs and sheep to influence the sheep's behavior. 

In our prior work \cite{grover2022noncooperative}\footnote{accepted in IEEE Conference on Decision and Control 2022}, we developed a centralized algorithm to solve this problem using control barrier functions. In this work, a) we provide more experimental validation of the centralized algorithm, b) propose two distributed algorithms, and c) provide simulations and experiments to validate these algorithms. Our formulation computes the velocity of each dog locally to prevent sheep from breaching the protected zone(s). In the first distributed algorithm, we allocate each sheep to a unique dog and pose a constraint on that dog's velocity to herd its allocated sheep away from the protected zone. We provide proof of feasibility of this approach, thus showing that whenever the number of sheep and dogs are equal, the herding problem is well-posed. Our previously proposed centralized algorithm lacked this feasibility guarantee. However, it did not necessitate equal numbers of dogs and sheep; in fact, in many experiments, fewer dogs than sheep were sufficient to herd all the sheep away. This observation led us to develop the second algorithm. In this algorithm, we construct an iterative distributed approach that asymptotically attains the same velocities as computed by the centralized approach, thereby attaining the same total optimality (measured in terms of the total movement the dogs exhibit) as the centralized approach and obviating the need to have equal numbers of dogs and sheep. We build on the dual-decomposition algorithms proposed in \cite{falsone2017dual,notarstefano2019distributed} for developing this distributed algorithm. Both of our proposed distributed algorithms are compositional in nature \textit{i.e.}, we can protect multiple zones by including more constraints, as shown in figure \ref{fig:dC0}. To highlight the performance of our formulation, we provide results from numerical simulations showing the success of our approach for multiple dogs against multiple sheep. Finally, we demonstrate our algorithm on real robots and show multiple dog robots successfully preventing the breaching of protected zones against multiple sheep robots.

The outline of this paper is as follows: in section \ref{priorwork}, we give a brief review of the prior work in this area. In section \ref{problem formulation},  we provide a mathematical formulation of the problem statement. In section \ref{controller design}, we show how to use control barrier functions to pose constraints on dog velocities. Section \ref{results} provides simulations and experimental results to demonstrate the proposed approach. Finally, we summarize our work in section \ref{conclusions} along with our directions for future work.
\section{Prior Work}
\label{priorwork}

The framework of multi-group interaction within MRS has many applications beyond the adversarial problem statements. The shepherding problem is an example of such a category. In \cite{lien2004shepherding,pierson2017controlling}, the authors have proposed methods to enable multiple shepherd agents to influence a flock of sheep by modeling the interaction as repulsion forces. The Robot Sheepdog Project \cite{vaughan1998robotA,vaughan2000experiments} conducted a real-world demonstration of a shepherding algorithm where a group of mobile ground robots cooperatively herded a flock of ducks to a given goal location.

In the literature, there are several works on non-cooperative shepherding as an example of a multi-group interaction type problem. The works like \cite{pierson2017controlling}, \cite{pierson2015bio}, \cite{licitra2017singleA}, \cite{licitra2017singleB}, \cite{sebastian2021multi}, \cite{bacon2012swarm}.  deal with a problem where the sheep robots do not exhibit adversarial behavior. They do not have any goals of their own. However, they experience a repulsive force from the dog robots, which is exploited to produce the desired behavior in the sheep robots. For example, collecting all the sheep at some location and then driving them to a target goal.

Differently from prior work, our sheep may or may not be adversarial. We call them adversarial if their goal lies inside the protected zone and non-adversarial otherwise. Our safe control synthesis approach remains the same regardless. The dog robots observe and generate their control commands considering the cohesion between the sheep robots, the attraction to their goal location, and the repulsion experienced by them from the dog robots. And as we use control barrier functions to generate the constraints on the velocity of the dog robots, it only requires the dynamics of the sheep to be represented as a symbolic function. Thus allowing for the sheep to experience any kind of attractive or repulsive forces. 

\section{Problem Formulation}

\label{problem formulation}
Consider a scenario with $m$ sheep agents flocking towards a common goal location. One commonly assumed model for flocking is the Reynolds-Boids dynamics \cite{10.1145/37401.37406} that considers inter-sheep cohesive forces, inter-sheep repulsive forces, and attraction to a common goal.
In the presence of \textit{dog agents}, each sheep's dynamics would include repulsive forces from each dog robot. While en route to their goal, the sheep, having no knowledge about high-value regions in workspace (\textit{protected zones}),  pose a risk of breaching them. Thus, our problem is to orchestrate the motions of dog robots by capitalizing on the repulsions that the sheep experience from the dogs to prevent this breaching. Next, we pose this problem in formal terms.

Consider the protected zone $\mathcal{P} \subset \mathbb{R}^2$ as a disc centered at $\boldsymbol{x}_P$ with radius $R_p$, \textit{i.e.}, $\mathcal{P} \coloneqq \{\boldsymbol{x} \in \mathbb{R}^2 \vert \ \ \|\boldsymbol{x}-\boldsymbol{x}_P\|\leq R_p\}$. We denote the flock of sheep as $\mathcal{S}$ and the position of the $i^{th}$ sheep as $\boldsymbol{x}_{S_i} \in \mathbb{R}^2$. The collective positions of all sheep is denoted as $\boldsymbol{x}^{all}_{S} \coloneqq (\boldsymbol{x}_{S_1},\boldsymbol{x}_{S_2},...,\boldsymbol{x}_{S_m})$. Similarly, we denote the set of all dogs using $\mathcal{D}$. The position of the $k^{th}$ dog is $\boldsymbol{x}_{D_k} \in \mathbb{R}^2$ and the positions of all dogs collectively is $\boldsymbol{x}^{all}_{D} \coloneqq (\boldsymbol{x}_{D_1},\boldsymbol{x}_{D_2},...,\boldsymbol{x}_{D_n})$. Each sheep follows single integrator dynamics $\Dot{\boldsymbol{x}}_{S_i} \coloneqq \boldsymbol{f}_i(\boldsymbol{x}_{S_1},...,\boldsymbol{x}_{S_n},\boldsymbol{x}_{D_1},...,\boldsymbol{x}_{D_n})$, given by 
\begin{align}
\label{sheepdynamics}
    \Dot{\boldsymbol{x}}_{S_i} =  \boldsymbol{u}_{S_i} &= k_{S} \underbrace{\sum_{j \in \mathcal{S}\backslash i}\left(1-\frac{R_{S}^{3}}{\|\boldsymbol{x}_{S_j} - \boldsymbol{x}_{S_i}\|^{3}}\right) (\boldsymbol{x}_{S_j} - \boldsymbol{x}_{S_i})}_{\mbox{inter-sheep cohesion and repulsion}} \nonumber  + \underbrace{k_{G} \left(\boldsymbol{x}_{G}-\boldsymbol{x}_{S_i}\right)}_{\mbox{attraction to goal}}
\\&+  \underbrace{k_{D} \sum_{l \in \mathcal{D}}  \frac{\boldsymbol{x}_{S_i} - \boldsymbol{x}_{D_l}}{\|\boldsymbol{x}_{S_i} - \boldsymbol{x}_{D_l}\|^{3}}}_{\mbox{repulsion from dogs}} 
\end{align}
Here, $R_S$ is a safety margin that each sheep tends to maintain with every other sheep, $\boldsymbol{x}_G$ is the sheep's desired goal and $k_S$, $k_G$ and $k_D$ are proportional gains corresponding to the attractive and repulsive forces.
We model each dog as a velocity controlled robot with the following dynamics:
\begin{align}
\label{doginput}
 \Dot{\boldsymbol{x}}_{D_k} = \boldsymbol{u}_{D_k} \hspace{0.2cm} \forall k \in \{1,2,\cdots,n\}
\end{align}
Before posing the problem, we state some assumptions on the dogs' knowledge:
\begin{assumption}
	\label{ass1}
	The dog robots have knowledge about the sheep's dynamics \textit{i.e.} \eqref{sheepdynamics} and can measure the sheep's positions accurately. 
\end{assumption}
\begin{assumption}
	\label{ass2}
	Each dog robot can measure the velocities of other dog robots (by using numerical differentiation, for example). 
\end{assumption}
\begin{mydef}
\label{problem1}
Assuming that the initial positions of the sheep $\boldsymbol{x}_{S_i}(0) \notin  \mathcal{P}$ $\forall i \in \mathcal{S}$, the dog robots' problem is to synthesize controls $\{\boldsymbol{u}_{D_1},\cdots, \boldsymbol{u}_{D_n}\}$ such that $\boldsymbol{x}_{S_i}(t) \notin  \mathcal{P}$ $\forall t\geq 0$ $\forall i \in \mathcal{S}$. 
\end{mydef}
\section{Controller Design}
\label{controller design}
In this section, we show two approaches to solve Problem \ref{problem1}, building on our previously proposed centralized algorithm \cite{grover2022noncooperative}. Define a safety index $h(\cdot):\mathbb{R}^2 \longrightarrow \mathbb{R}$ that quantifies the distance of $S_i$ from $\mathcal{P}$: 
\begin{align}
\label{hdef}
    h(\boldsymbol{x}_{S_i}) = \|\boldsymbol{x}_{S_i} - \boldsymbol{x}_P\|^2 - (r+R_p)^2
\end{align}
Here $r$ is a safety buffer distance. Thus, we require $h(\boldsymbol{x}_{S_i}(t)) \geq 0$ $\forall t \geq 0$.  We define $\boldsymbol{x}=(\boldsymbol{x}^{all}_S,\boldsymbol{x}^{all}_D)$ as the aggregated state of all sheep and all dogs. To ensure, $h(\boldsymbol{x}_{S_i}(t)) \geq 0$ $\forall t \geq 0$, we treat $h(\cdot)$ as a control barrier function require its derivative to satisfy

\begin{align}
\label{hdot}
    \Dot{h}(\boldsymbol{x}) + p_1h(\boldsymbol{x}_{S_i}) \geq 0.  
\end{align}
Here $p_1$ is a design parameter and is chosen such that it satisfies:
\begin{align}
    \label{eq:condition_on_p1}
    p_1 > 0 \quad \text{and} \quad p_1 > -\frac{\Dot{h}(\boldsymbol{x}(0))}{h(\boldsymbol{x}_{S_i}(0))}.
\end{align}
\noindent The first condition on $p_1$ ensures the pole is real and negative. The second depends on the initial positions $\boldsymbol{x}(0)$ of all the sheep and dogs relative to the protected zone. Note that the constraint in \eqref{hdot} does not contain any dog velocity terms, which is what we require to control each dog. Therefore, we define the LHS of \eqref{hdot} as another control barrier function $v(\boldsymbol{x}):\mathbb{R}^{4n} \longrightarrow \mathbb{R}$:
\begin{align}
\label{vdef}
    v = \Dot{h} + p_1h,
\end{align}
and require its derivative to satisfy the constraint:  
$\Dot{v}(\boldsymbol{x}) + p_2v(\boldsymbol{x}) \geq 0.$
Here $p_2$ is another design parameter which must satisfy
\begin{align}
\label{eq:condition_on_p2}
    p_2 > 0 \quad \text{and} \quad p_2 > -\frac{\Ddot{h}(\boldsymbol{x}(0)) + p_1\Dot{h}(\boldsymbol{x}(0))}{\Dot{h}(\boldsymbol{x}(0)) + p_1h(\boldsymbol{x}_{S_i}(0))}
\end{align}
Using \eqref{hdef}, \eqref{vdef} and the constraint on the derivative, 
 we get
\begin{align}
\label{timederivatives}
\Ddot{h}(\boldsymbol{x}) + \alpha \Dot{h}(\boldsymbol{x}) + \beta h(\boldsymbol{x}_{S_i}) \geq 0 
\end{align}
where $\alpha \coloneqq p_1 +p_2$ and $\beta \coloneqq p_1p_2$. The derivatives of $h(\cdot)$ are:
\begin{align}
\label{hd}
    \Dot{h}(\boldsymbol{x}) &= 2(\boldsymbol{x}_{S_i} - \boldsymbol{x}_P)^T\Dot{\boldsymbol{x}}_{S_i} 
    =2(\boldsymbol{x}_{S_i} - \boldsymbol{x}_P)^T \boldsymbol{f}_i(\boldsymbol{x}) \\
    \label{hdd}
    \Ddot{h}(\boldsymbol{x}) &= 2 \boldsymbol{f}^T_{i} \boldsymbol{f}_{i}    + 2(\boldsymbol{x}_{S_i}-\boldsymbol{x}_P)^T\bigg(\sum_{j \in \mathcal{S}}\mathbb{J}_{ji}^S \boldsymbol{f}_{i} + \sum_{l \in \mathcal{D}}\mathbb{J}_{li}^D\boldsymbol{u}_{D_l}\bigg)
\end{align}
\noindent where the jacobians are defined as $\mathbb{J}^S_{ji} \coloneqq \nabla_{\boldsymbol{x}_{S_j}} \boldsymbol{f}_{i} (\boldsymbol{x})$  and $\mathbb{J}^D_{li} \coloneqq \nabla_{\boldsymbol{x}_{D_l}} \boldsymbol{f}_{i} (\boldsymbol{x})$
Note that  \eqref{hdd} contains the velocity terms of all dogs. In \cite{grover2022noncooperative}, we leveraged this observation to obtain a linear constraint on the velocity of all dogs collectively for preventing sheep $S_i$ from breaching $\mathcal{P}$:
\begin{align}
    \label{herdingcon2}
    A^C_i \boldsymbol{u}^{all}_D\leq b^C_i, \hspace{0.5cm} \mbox{where}
\end{align}
\vspace{-5.5ex}
\begin{align*}
    A^{C}_{i} &\coloneqq  (\boldsymbol{x}_P -\boldsymbol{x}_{S_i})^T
    \begin{bmatrix}
     \mathbb{J}_{1i}^D, \hspace{0.1cm} \mathbb{J}_{2i}^D, \hspace{0.1cm}\cdots,\hspace{0.1cm}\mathbb{J}_{ni}^D
    \end{bmatrix}\\
      b^{C}_{i} &\coloneqq\boldsymbol{f}^T_i\boldsymbol{f}_i
      + (\boldsymbol{x}_{S_i} - \boldsymbol{x}_P)^T\sum_{j \in \mathcal{S}}\mathbb{J}_{ji}^S \boldsymbol{f}_j+ \alpha(\boldsymbol{x}_{S_i}-\boldsymbol{x}_P)^T\boldsymbol{f}_i + \beta\frac{h}{2} 
\end{align*}
This gives us a centralized algorithm  that collectively computes the velocities of all dogs using the following QP:
\begin{align}
\label{centralized_dog_control}
    \boldsymbol{u}^{all}_{D} = \underset{\boldsymbol{u}^{all}_{D}}{\arg \min }\|\boldsymbol{u}^{all}_{D}\|^{2} \nonumber \\
    \text{subject to} \quad A^C_i\boldsymbol{u}^{all}_{C} \leq \boldsymbol{b}^C_i  \hspace{0.1cm}\forall i \in \mathcal{S}.
\end{align}
Building on this centralized approach, in this paper, we develop two distributed approaches wherein we allow each dog to compute its velocity locally. The computed velocities will make the dog herd the sheep away from  $\mathcal{P}$. 
\subsection{Approach 1: One dog to one sheep allocation based approach}
\label{section:4.1}
In this approach, we assume that we have an equal number of dogs and sheep. By exploiting this equality, we assign a unique sheep $S_i$ for $i \in \{1,\cdots,n\}$ to a unique dog $D_k$ for $k \in \{1,\cdots,n\}$ and make $D_k$ responsible for herding $S_i$ away from $\mathcal{P}$. In other words, $D_k$ computes a velocity $\boldsymbol{u}_{D_k}$ that repels $S_i$ from $\mathcal{P}$ thereby ensuring that $\boldsymbol{x}_{S_i}(t) \notin  \mathcal{P}$ $\forall t\geq 0$. The premise is that owing to the equality, each sheep will end up being herded by a unique dog, therefore, no sheep will breach the protected zone \footnote{Note that although $S_i$ is assigned to $D_k$, the position of the remaining dogs $\{1,\cdots,n\}\backslash k$ and the remaining sheep $\{1,\cdots,n\}\backslash i$ do influence $D_k$'s constraint parameters ($A^H_i,b^H_i$), and in turn, its computed velocity $\boldsymbol{u}^*_{D_k}$.}. Now while this strategy necessitates having an equal number of dogs and sheep, the benefit of this approach stems from the feasibility guarantee (that we prove shortly), which the centralized approach lacks. Simple algebraic manipulation of constraint \eqref{herdingcon2} yields a constraint on the velocity of $D_k$ as follows
\begin{align}
    \label{herdingcon1}
    A^H_i \boldsymbol{u}_{D_k}\leq b^H_i, \hspace{0.5cm} \mbox{where}
\end{align}
\vspace{-4.5ex}
\begin{align*}
    A^H_{i} &\coloneqq  (\boldsymbol{x}_P -\boldsymbol{x}_{S_i})^T
     \mathbb{J}_{ki}^D\\
      b^H_{i} &\coloneqq\boldsymbol{f}^T_i\boldsymbol{f}_i
       + \beta\frac{h}{2} + (\boldsymbol{x}_{S_i} - \boldsymbol{x}_P)^T \Bigl\{\sum_{j \in \mathcal{S}}\mathbb{J}_{ji}^S \boldsymbol{f}_j+ \alpha \boldsymbol{f}_i + \sum_{l \in \mathcal{D}\backslash k} \mathbb{J}_{li}^D \boldsymbol{u}_{D_l} \Bigr\}
\end{align*}
Here $A^H_i \in \mathbb{R}^{1 \times 2}$  and $\boldsymbol{b}^H_i \in \mathbb{R}$. The term $u_{D_l}$ in the expression of $b^H_{i}$ is computed by using numerical differentiation of the positions $\boldsymbol{x}_{D_l}$. We pose a QP to obtain the min-norm velocity for $D_k$ as follows
\begin{align}
\label{dog_control}
    \boldsymbol{u}^{*}_{D_k} = \underset{\boldsymbol{u}_{D_k}}{\arg \min }\|\boldsymbol{u}_{D_k}\|^{2} \nonumber \\
    \text{subject to} \quad A^H_i\boldsymbol{u}_{D_k} \leq b^H_i  
\end{align}
The obtained velocity  $\boldsymbol{u}^{*}_{D_k}$ guarantees that the protected zone $\mathcal{P}$ will not be breached by sheep $S_i$ by ensuring that $h(\boldsymbol{x}_{S_i}(t)) \geq 0$ $\forall t \geq 0$. Since each dog in $\mathcal{D}$ is in-charge of herding exactly one sheep in $\mathcal{S}$, feasibility of \eqref{herdingcon1} $\forall k \in \mathcal{D}$ would ensure no sheep breaches $\mathcal{P}$. Next, we show the conditions under which \eqref{dog_control} remains feasible but first state some assumptions.
\begin{assumption}
\label{ass4} We make the following assumptions on the distances between pairs of agents:
\begin{enumerate}
    \item There exists a lower bound and upper bound on the distance between any pair of sheep, i.e, $L_S \leqslant \left\|\boldsymbol{x}_{S_{i}}-\boldsymbol{x}_{S_{j}}\right\| \leqslant M_{S}$, $\forall i,j \in \mathcal{S}$ and $i \neq j$.
    \item  There exists a lower bound on the distance between every sheep and dog, i.e., $ \left\|\boldsymbol{x}_{S_{i}}-\boldsymbol{x}_{D_{k}}\right\| \geq L_{D}$ $\forall i \in \mathcal{S}$ and $k \in \mathcal{D}$.
    \item  There exists a upper bound on the distance between each sheep and its goal \textit{i.e.}, $ \left\|\boldsymbol{x}_{S_{i}}-\boldsymbol{x}_{G}\right\| \leqslant M_{G}$ and  between the sheep and the center of the protected zone \textit{i.e.}, $\left\|\boldsymbol{x}_{S_{i}}-\boldsymbol{x}_P\right\| \leqslant M_{P}$.
\end{enumerate}
\end{assumption}
\begin{theorem}
\label{theorem_caseA} In a scenario with `$n$' dogs and `$n$' sheep, with each dog assigned a unique sheep, the herding constraint \eqref{herdingcon1} for a given dog is always feasible, provided assumptions \ref{ass4} are met.
\end{theorem}
\vspace{-2.5ex}
\begin{proof}
See appendix (section \ref{dist_appendix}).
\end{proof}
\subsection{Approach 2: Iterative distributed reformulation of \eqref{centralized_dog_control} }
\label{section:4.2}
The distributed formulation proposed in \eqref{dog_control} comes with a feasibility guarantee ensuring that all sheep will be herded away from $\mathcal{P}$. While vital, this comes at the cost of requiring as many dog robots as the number of sheep agents. This is because, in a way, this equality ensures that controlling the sheep from the perspective of dog robots is not an underactuated problem. Be that as it may, in our simulations and experiments involving the centralized approach with an equal number of dogs and sheep, we frequently observed that not all dog robots needed to move to repel the sheep away from $\mathcal{P} $ \textit{i.e.,} equality may have been an overkill. Thus, in terms of budget efficiency, at least empirically, the centralized approach outweighs the distributed approach. 

This raises the question, can we convert the centralized algorithm of \eqref{centralized_dog_control} into a distributed version that inherits the budget efficiency (optimality) promised by \eqref{centralized_dog_control}? Indeed, we found out that \cite{falsone2017dual,notarstefano2019distributed} propose algorithms to convert constrained-coupled convex optimization  problems (such as \eqref{centralized_dog_control}) into  distributed counterparts. They combine techniques called dual decomposition and proximal minimization and develop iterative distributed schemes which consist of local optimization problems. The solutions to these optimization problems asymptotically converge to the solution of centralized optimization under mild convexity assumptions and connectivity properties of the communication network. In our case, this network refers to the communication between dog robots. Below, we present the distributed dual sub-gradient method of \cite{falsone2017dual,notarstefano2019distributed} adapted to the costs and constraints of \eqref{centralized_dog_control}. This algorithm calculates an estimate of dog $D_k$'s velocity $\boldsymbol{\hat{u}}_{D_k}$ which, given large enough iterations $K_{max}$, matches with the $k^{th}$ velocity component in the optimal velocities $\boldsymbol{u}^{*all}_D$ returned by \eqref{centralized_dog_control}. $A_k \in \mathbb{R}^{n_S \times 2}$ refers to those columns of $A^H$ that correspond to $\boldsymbol{u}_{D_k}$ in $\boldsymbol{u}^{all}_D$.
\begin{algorithm}

\caption{Distributed Dual Subgradient for \eqref{centralized_dog_control} (based on sec. 3.4.2 in \cite{notarstefano2019distributed})}
\begin{algorithmic}[1]
\label{alg1}
      \Statex \hspace{-0.7cm} \textbf{Initialize Lagrange Multiplier}: $\boldsymbol{\mu}^0_k = \boldsymbol{0} \in \mathbb{R}^{n_S}$
     \Statex \hspace{-0.7cm}  \textbf{Evolution}: $t=1,2,\cdots,K_{max}$
    \Statex  \textbf{Gather Multipliers} $\boldsymbol{\mu}^t_r$ from $D_r$ $\forall r \in \{1,\cdots,n_D\}\backslash k$  
    \Statex  \textbf{Average Multipliers}: $\boldsymbol{v}^{t+1}_k = \frac{1}{n_D}\sum_{r \in \{1,\cdots,n_D\}\backslash k}\boldsymbol{\mu}^t_r$  
      \Statex  \textbf{Local Solution}: $\boldsymbol{u}^{t+1}_{D_k}= \underset{\boldsymbol{u}}{\arg \min}\norm{\boldsymbol{u}}^2 +(\boldsymbol{v}^{t+1}_k)^T(A_{k}\boldsymbol{u}-\frac{1}{n_D}\boldsymbol{b}^H)=-\frac{1}{2}A_{k}^T\boldsymbol{v}^{t+1}_k$   
      \Statex \textbf{Update Multiplier}: $\boldsymbol{\mu}^{t+1}_{k} = \big[\boldsymbol{v}^{t+1}_k + \gamma_t \big(A_{k}\boldsymbol{u}^{t+1}_{D_k} - \frac{1}{n_D}\boldsymbol{b}\big)\big]_{+}$ 
      \Statex \hspace{-0.7cm} \textbf{Return Average}: $\boldsymbol{\hat{u}}_{D_k} = (1/K_{max}) \sum_{t=1}^{K_{max}}\boldsymbol{u}^{t}_{D_k} $ 
\end{algorithmic}
\end{algorithm}

\section{Results}
\label{results}

In this section, we provide simulation and real-world experimental results demonstrating our proposed distributed algorithms. 

\subsection{Simulation Results}

\begin{figure*}
	\centering     
	\subfigure[Two dogs v. two sheep. ]{\label{fig:dA0}\includegraphics[trim={1.1cm 1.5cm 1.5cm 1.5cm},clip,width=0.32\columnwidth]{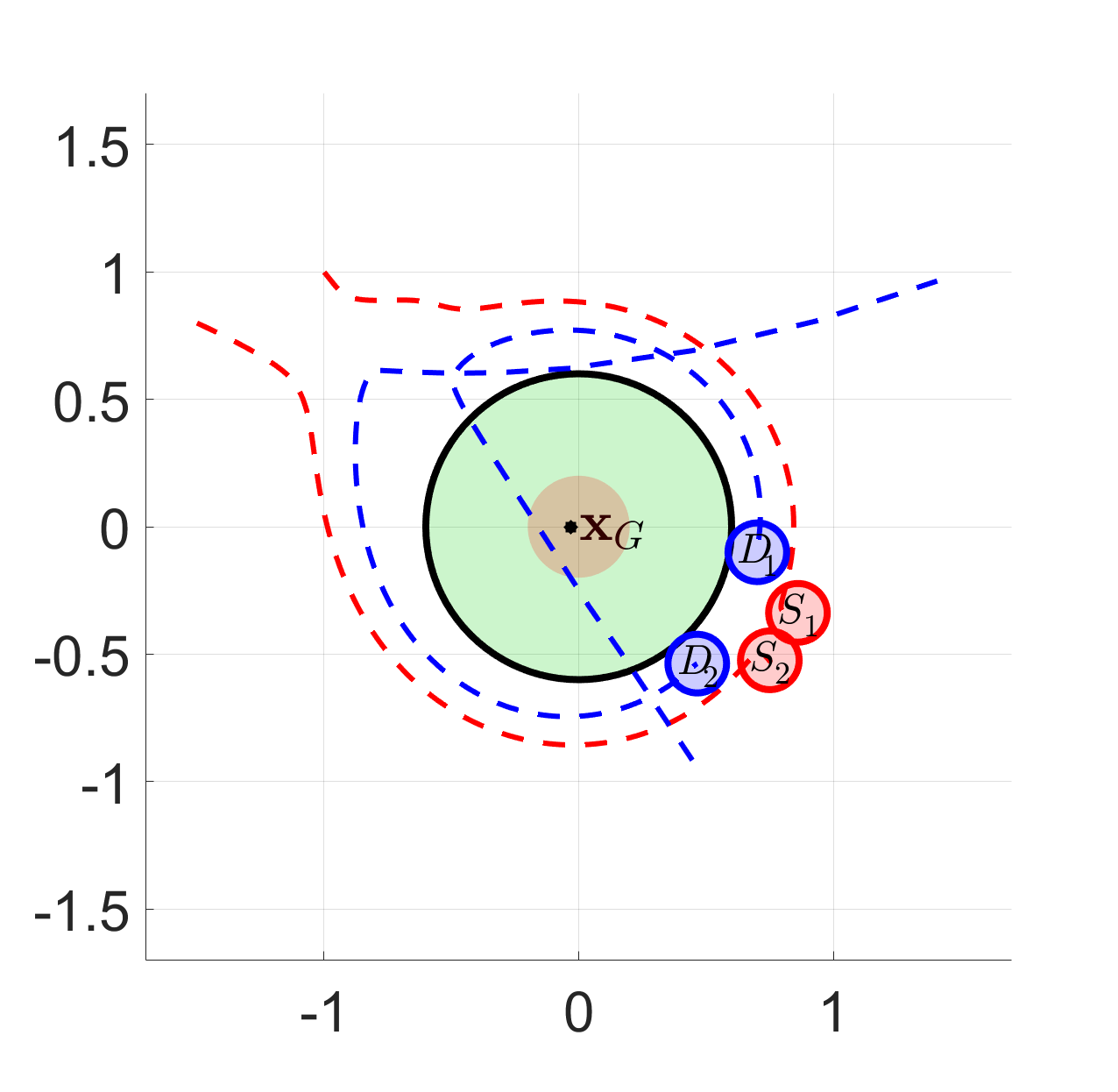}}
	\subfigure[\small Three dogs v. three sheep ]{\label{fig:dB0}\includegraphics[trim={1.1cm 1.5cm 1.5cm 1.5cm},clip,width=0.34\columnwidth]{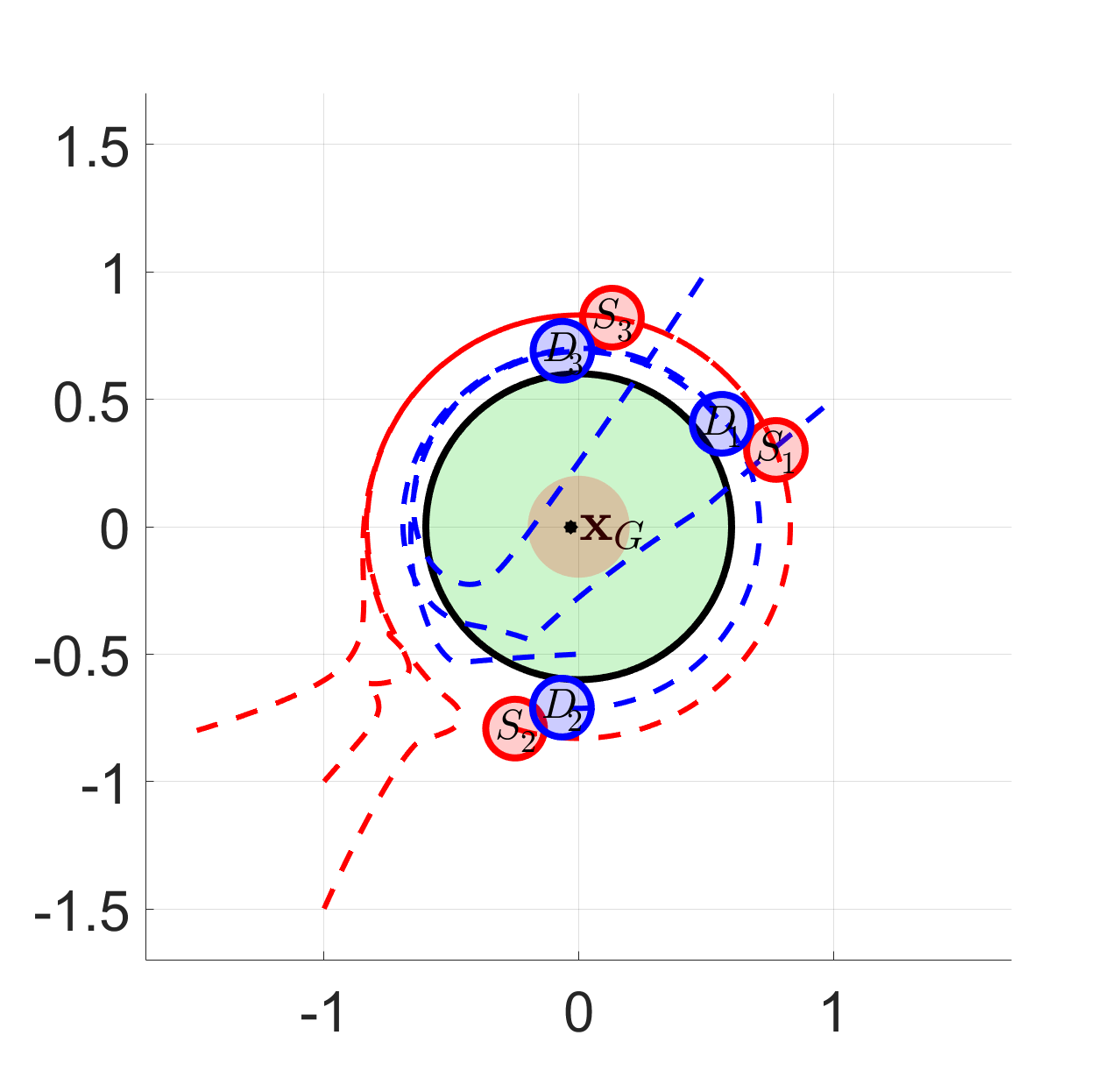}}
	\subfigure[Four dogs v. four sheep. ]{\label{fig:dC0}\includegraphics[trim={1.1cm 1.5cm 1.5cm 1.5cm},clip,width=.32\columnwidth]{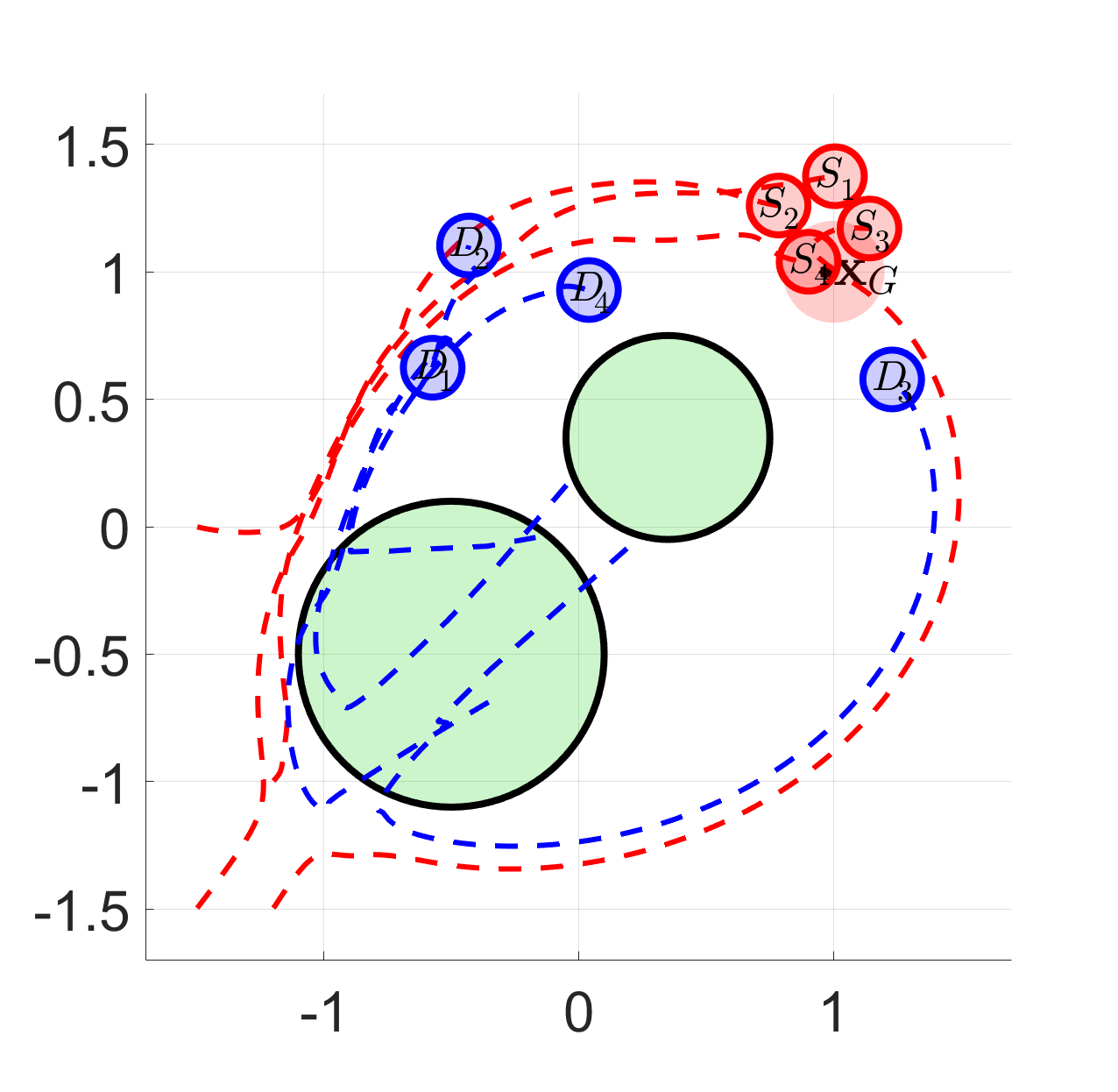}}
	\caption{Preventing the breaching of the protected zone using our proposed distributed algorithm in section \ref{section:4.1}. Here dogs are shown in blue and  sheep in red. The green disc represents the protected zone. The nominal task of the sheep is to go straight towards goal $\boldsymbol{x}_G$. However, since this would result in infiltration of the protected zone, the dog intervenes using the control algorithm presented in \eqref{dog_control}. In Fig. \ref{fig:dC0}, we defend two protected zones from four sheep.}
	\label{fig:defendingprotectedzon}
\end{figure*}

We first validate the first distributed algorithm and the feasibility proof given in \ref{section:4.1}. For this, we model the sheep with the Reynolds-Boids dynamics \eqref{sheepdynamics} with gains $k_S=0.5$, $k_G=1$ and $k_D=0.1$. The dogs use \eqref{dog_control} to compute their velocities, where hyperparameters $\alpha$ and $\beta$ are computed following \eqref{eq:condition_on_p1} and \eqref{eq:condition_on_p2}. We chose a circular protected zone of radius $R_p = 0.6$m and center $\boldsymbol{x}_P$ at origin. The sheep are initialized outside of the protected zone, and their goal location $\boldsymbol{x}_G$ is chosen such that their nominal trajectory would make them breach the zone, thus necessitating intervention from dogs. The positions of dogs are initialized randomly within a certain range of the protected zone. In figures \ref{fig:dA0} and \ref{fig:dB0}, we show two examples involving a) two dog robots vs. two sheep robots and b) three dog robots vs. three sheep robots. To demonstrate the compositionality of our approach, we consider two protected zones in figure \ref{fig:dC0} where we have four dogs defending both zones from four sheep. In all these simulations, none of the sheep breach any zone, thus demonstrating the correctness of our approach. In the interest of space, we skip the simulation results for the algorithm in \ref{section:4.2} but do provide experimental results.

\subsection{Robot Experiments}


In this section, we show the results obtained by performing robot experiments by implementing the distributed algorithms of section \ref{section:4.1} and section \ref{section:4.2}. Additionally, we also present more experimental results for our prior centralized algorithm from \cite{grover2022noncooperative} (because at the time, we did not have as many robots). We conduct these experiments in our lab's multi-robot arena, which consists of a 14ft $\times$ 7ft platform with multiple Khepera IV robots and eight Vicon cameras for motion tracking. Although Khepera robots have unicycle dynamics, \cite{grover2022noncooperative} consists of a  technique to convert the single-integrator dynamics (assumed for dogs and sheep) to linear and angular velocity commands for the robots. 

\begin{figure*}
	\centering     
	\subfigure[$t = 0s$ ]{\label{fig:dA3}\includegraphics[trim={0.4cm 0.1cm 0.1cm 0cm},clip,width=0.465\columnwidth]{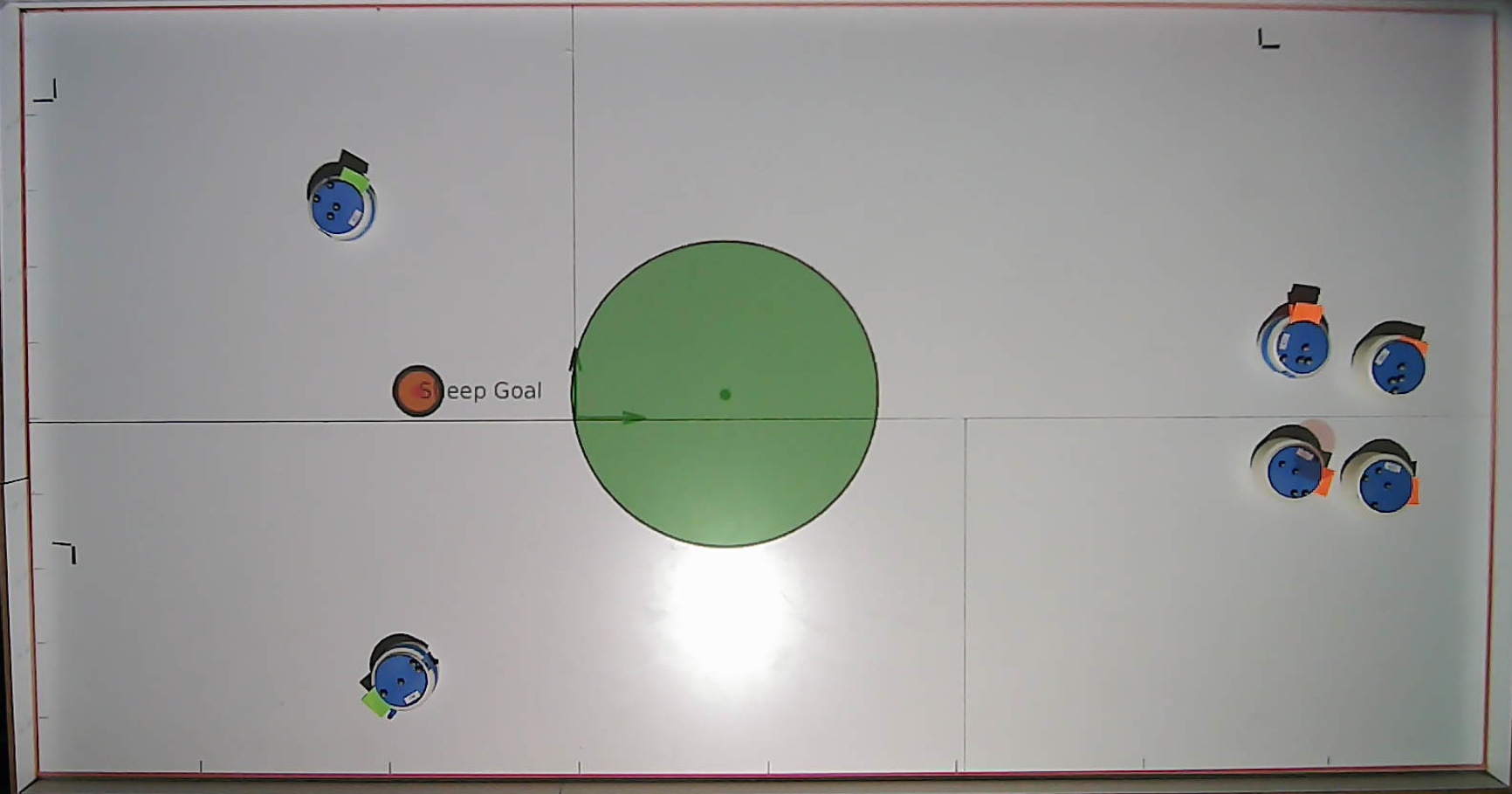}}
	\subfigure[$t = 5s$ ]{\label{fig:dB3}\includegraphics[trim={0.4cm 0.1cm 0.1cm 0cm},clip,width=0.465\columnwidth]{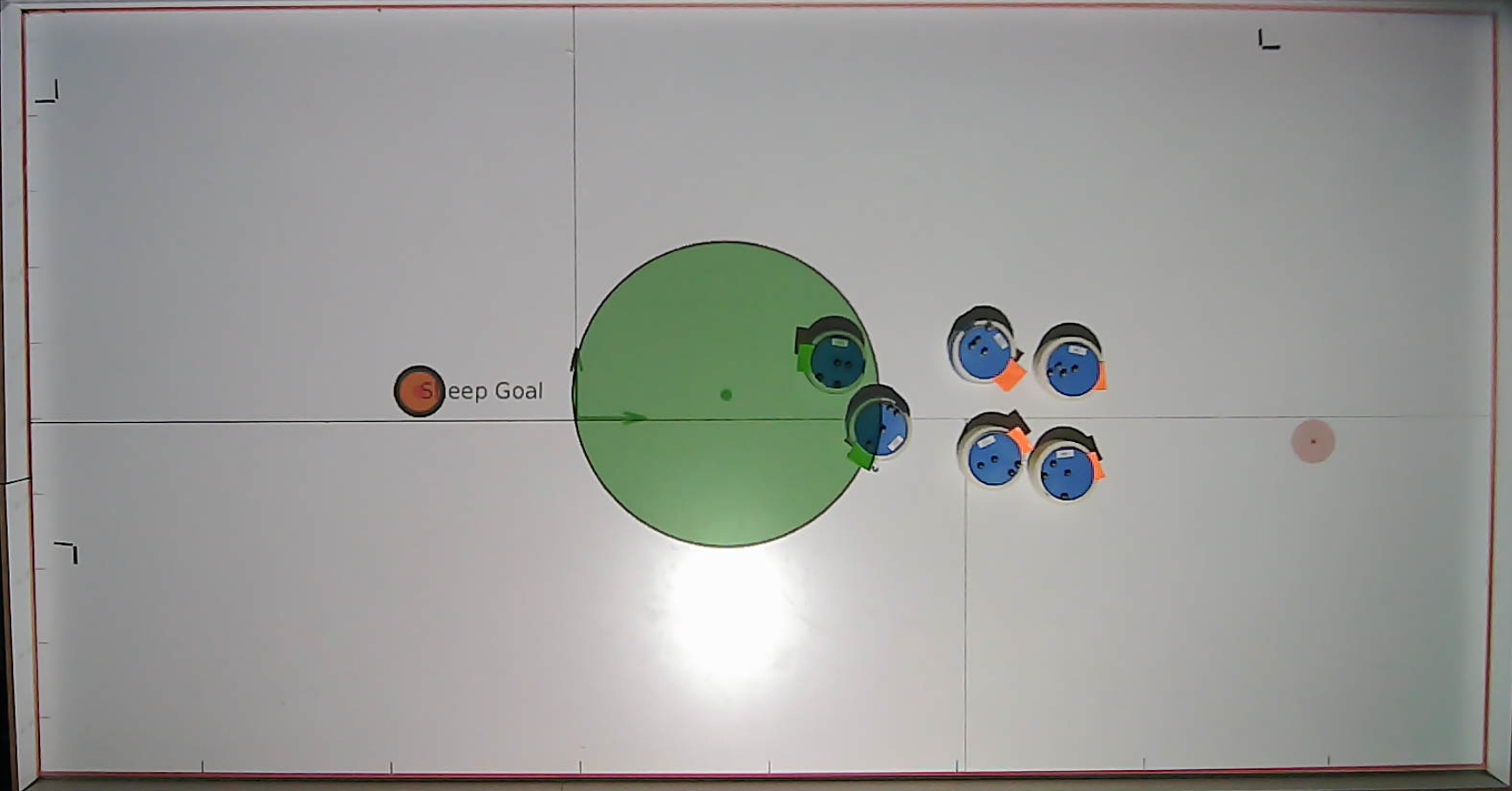}}
	\subfigure[$t = 12s$ ]{\label{fig:dC3}\includegraphics[trim={0.4cm 0.1cm 0.1cm 0cm},clip,width=0.465\columnwidth]{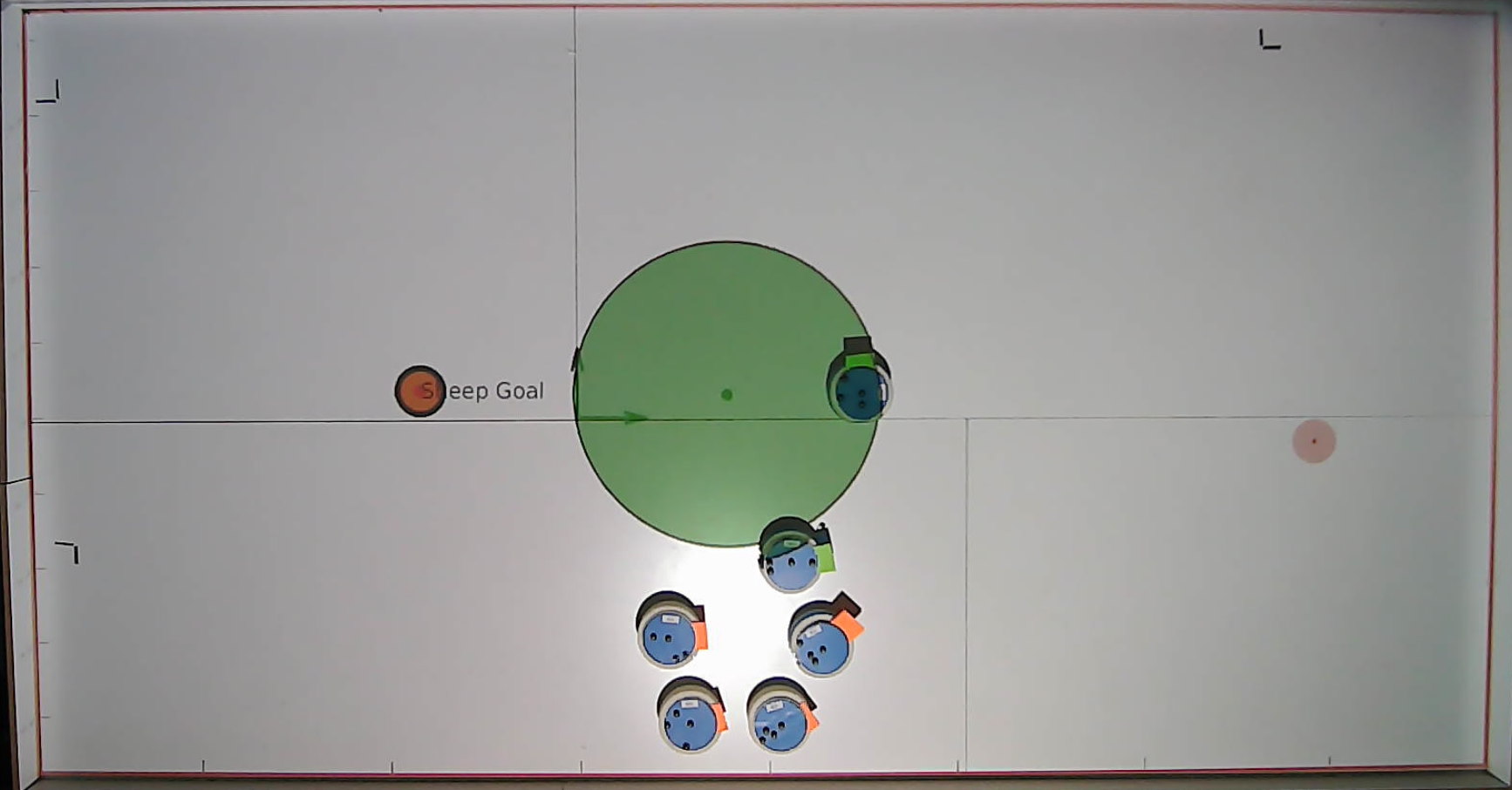}}
	\subfigure[$t = 30s$ ]{\label{fig:dC3}\includegraphics[trim={0.4cm 0.1cm 0.1cm 0cm},clip,width=0.465\columnwidth]{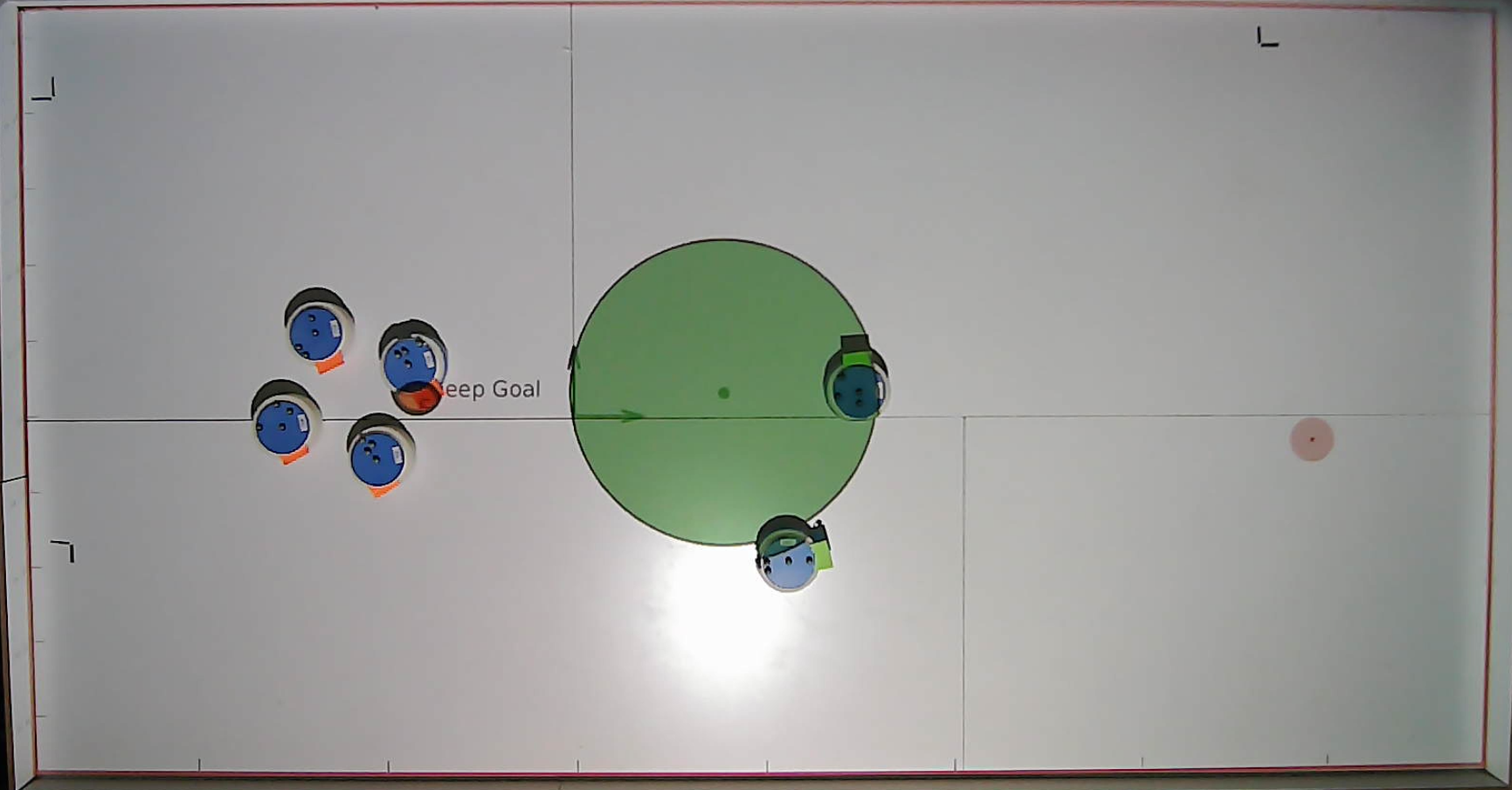}}
	\caption{\textbf{Experiments for Centralized Control:} Two dogs defending the protected zone from four sheep using centralized control algorithm \eqref{centralized_dog_control} from our prior work \cite{grover2022noncooperative}. Video at \url{https://bit.ly/3OTAnOu}.}
	\label{fig:2v4Cen}


	\centering     
	\subfigure[$t = 0s$ ]{\label{fig:dA3}\includegraphics[trim={0.4cm 0.1cm 0.1cm 0cm},clip,width=0.465\columnwidth]{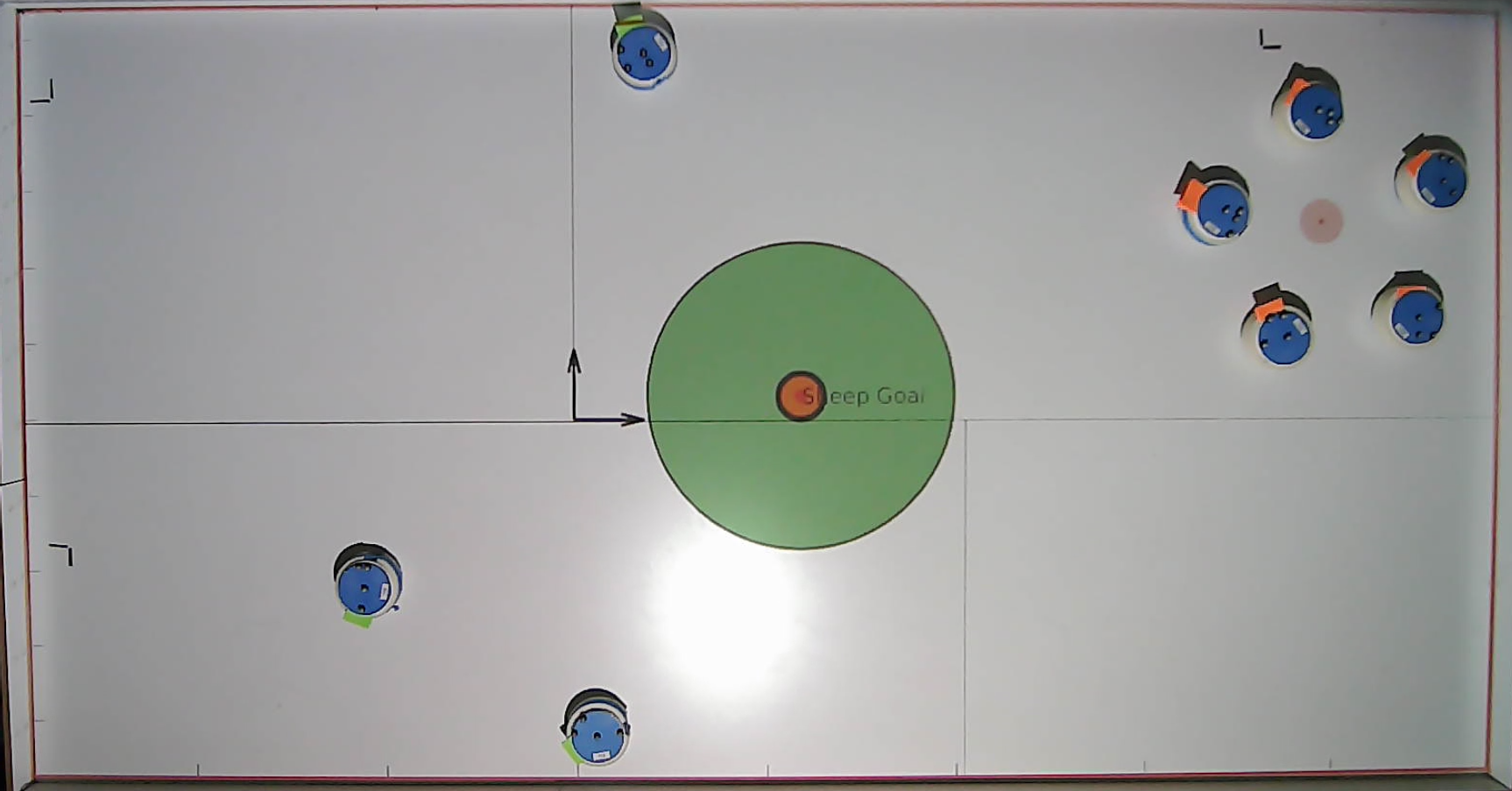}}
	\subfigure[$t = 5s$ ]{\label{fig:dB3}\includegraphics[trim={0.4cm 0.1cm 0.1cm 0cm},clip,width=0.465\columnwidth]{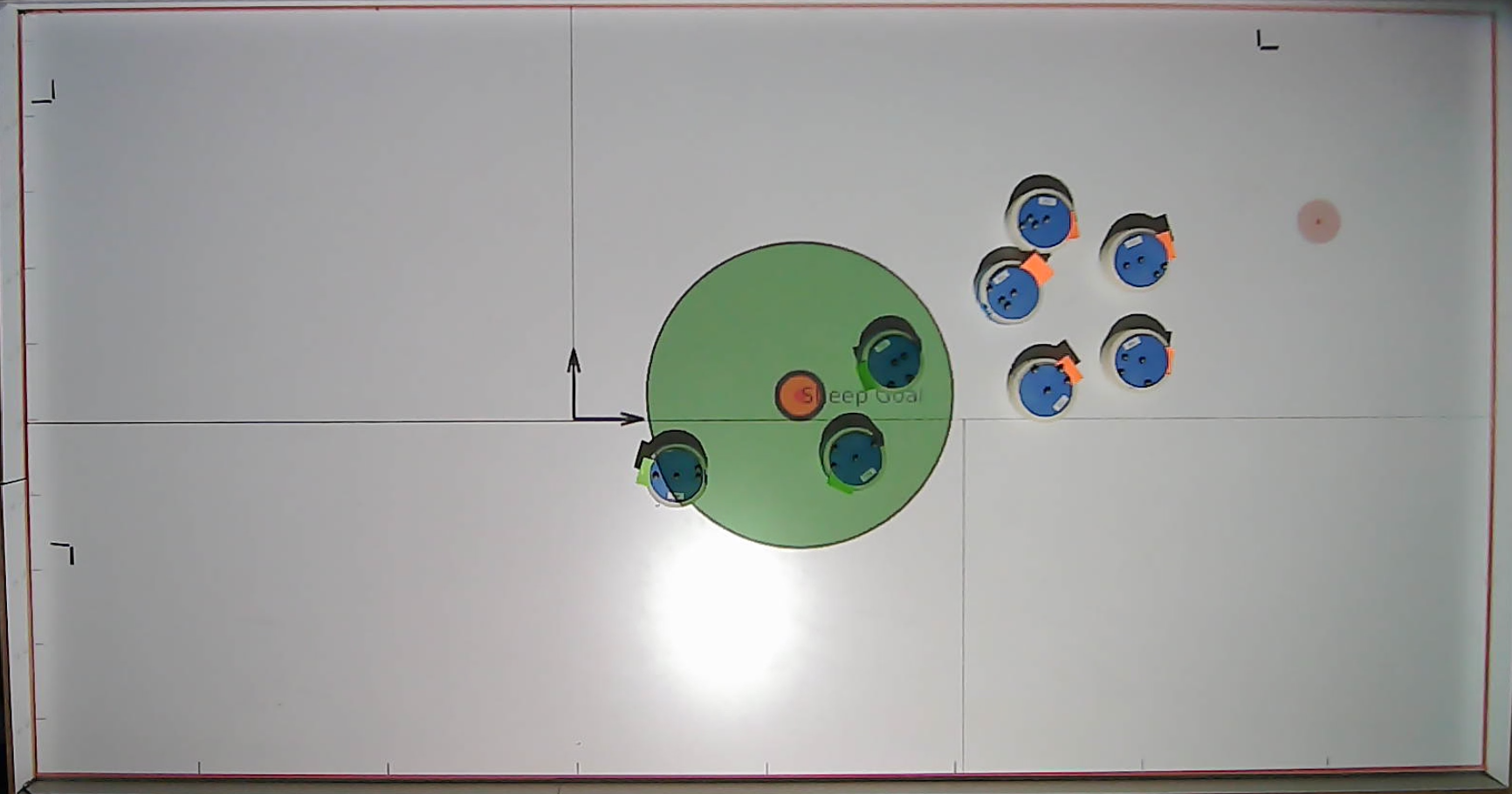}}
	\subfigure[$t = 30s$ ]{\label{fig:dC3}\includegraphics[trim={0.4cm 0.1cm 0.1cm 0cm},clip,width=0.465\columnwidth]{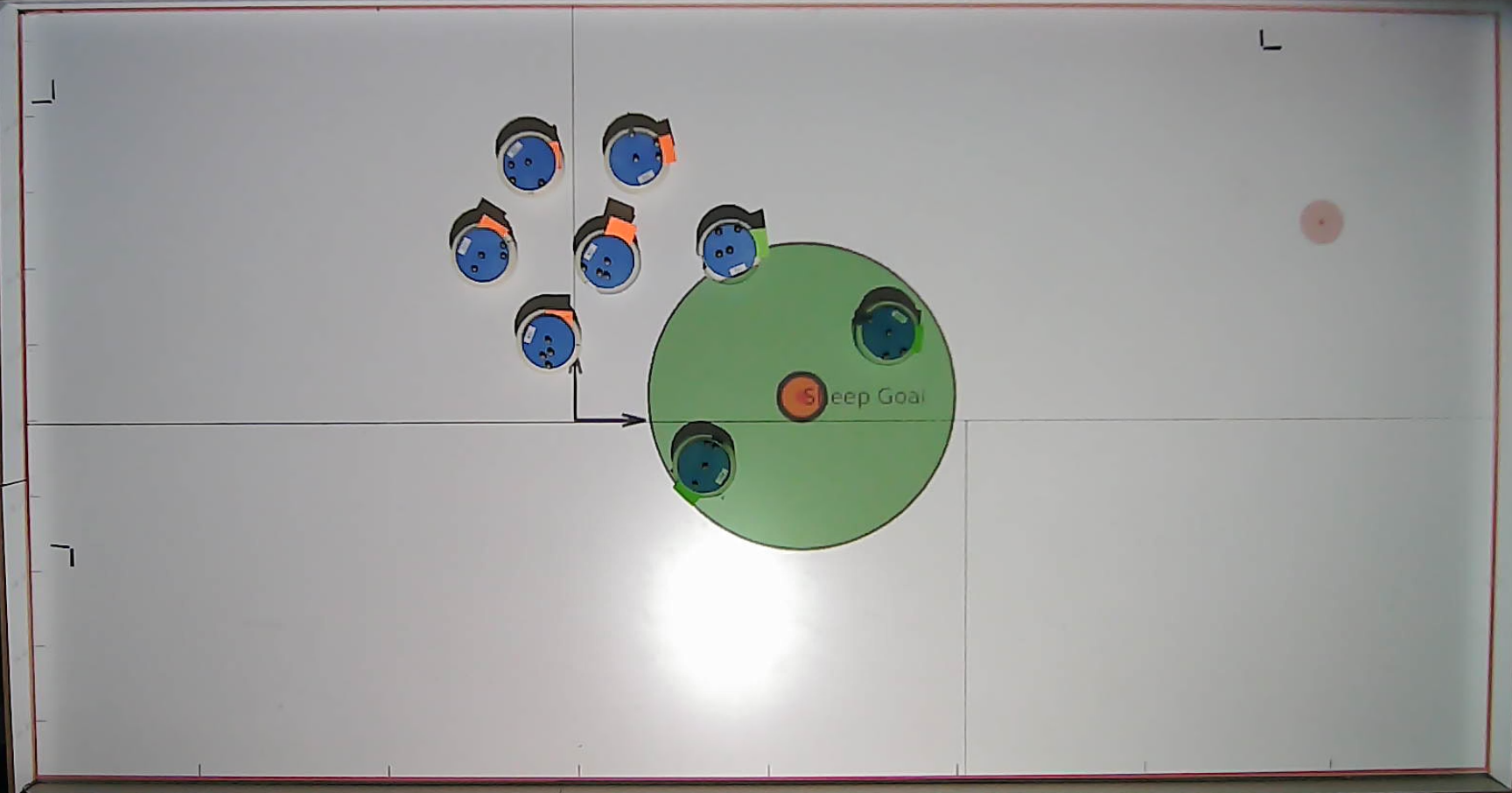}}
	\subfigure[$t = 50s$ ]{\label{fig:dC3}\includegraphics[trim={0.4cm 0.1cm 0.1cm 0cm},clip,width=0.465\columnwidth]{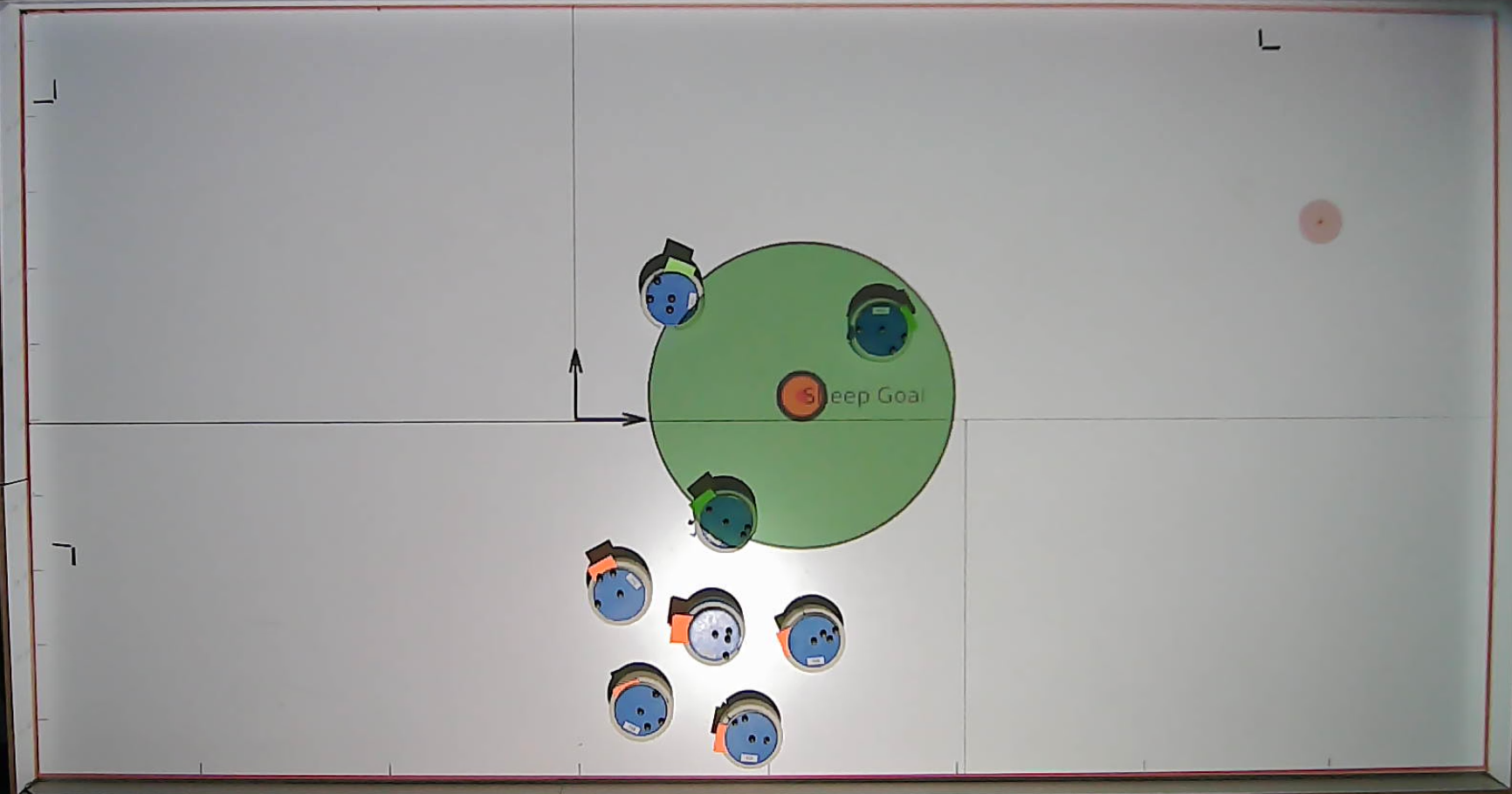}}
	\caption{\textbf{Experiment for Centralized Control:} Three dogs (green-tailed robots) defending a protected zone from five sheep (orange-tailed robots) using centralized control  \eqref{centralized_dog_control} from our prior work \cite{grover2022noncooperative}. Video at \url{https://youtu.be/2_Xuxnd9jZw}. }
	\label{fig:3v5Cen}
\end{figure*}

First of all, to build upon our previous work, we show additional experiments using centralized velocity computation of the dog robots \eqref{centralized_dog_control}. Figure \ref{fig:2v4Cen} shows a case with 2 dog and 4 sheep robots. The dog robots have a green tail, and the sheep robots have an orange tail. The tails are pointing in the opposite direction of the robot's heading angle. The protected zone is the green-colored circular region. This figure shows the performance in the case of an underactuated system, i.e, there are more sheep against less number of dogs. Another example is shown in figure \ref{fig:3v5Cen} where 3 dogs successfully prevent breaching against 5 sheep robots.

\begin{figure*}
	\centering     
	\subfigure[$t = 0s$ ]{\label{fig:dA3}\includegraphics[trim={0.4cm 0.1cm 0.1cm 0cm},clip,width=0.465\columnwidth]{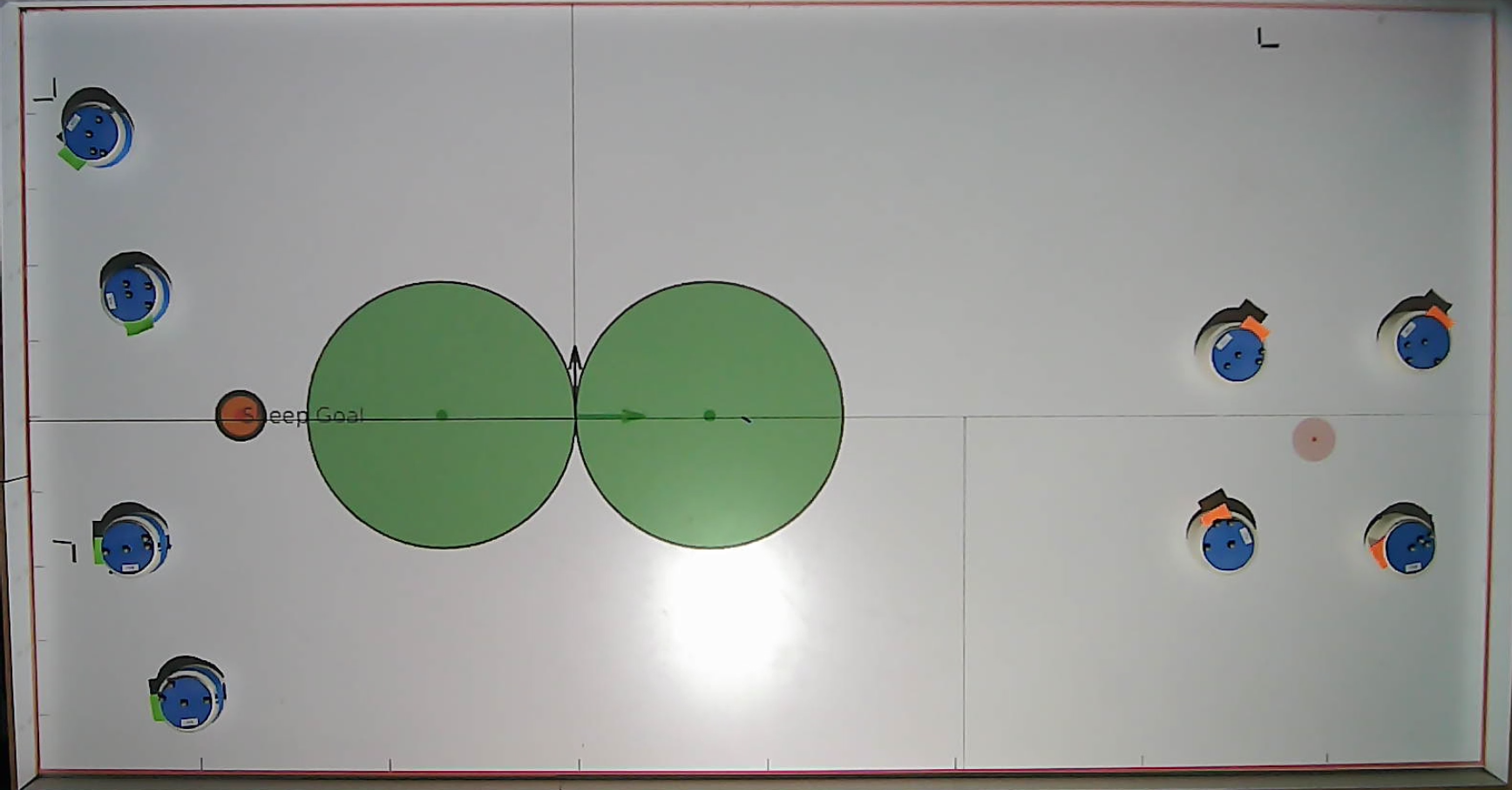}}
	\subfigure[$t = 6s$ ]{\label{fig:dB3}\includegraphics[trim={0.4cm 0.1cm 0.1cm 0cm},clip,width=0.465\columnwidth]{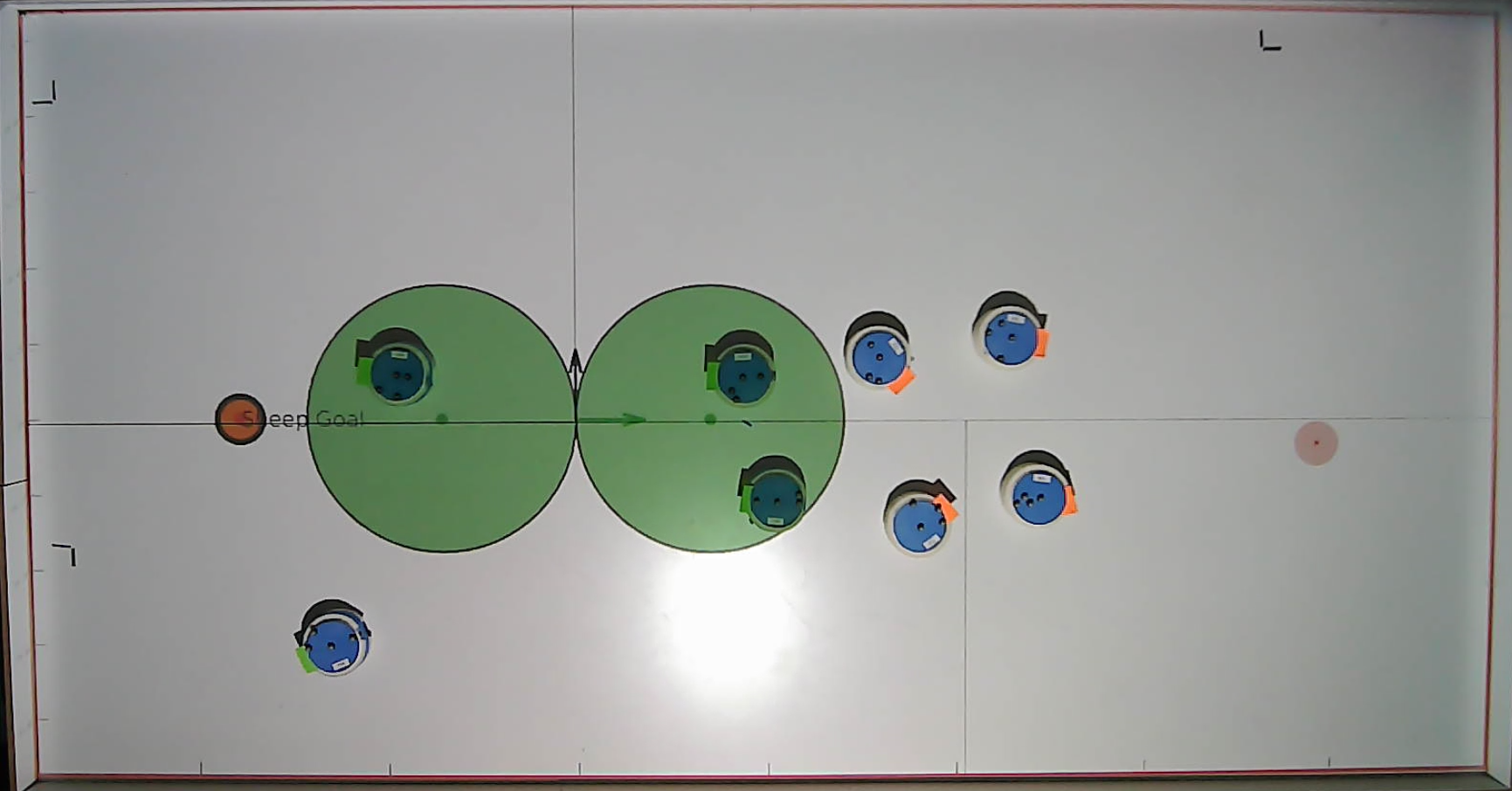}}
	\subfigure[$t = 12s$ ]{\label{fig:dC3}\includegraphics[trim={0.4cm 0.1cm 0.1cm 0cm},clip,width=0.465\columnwidth]{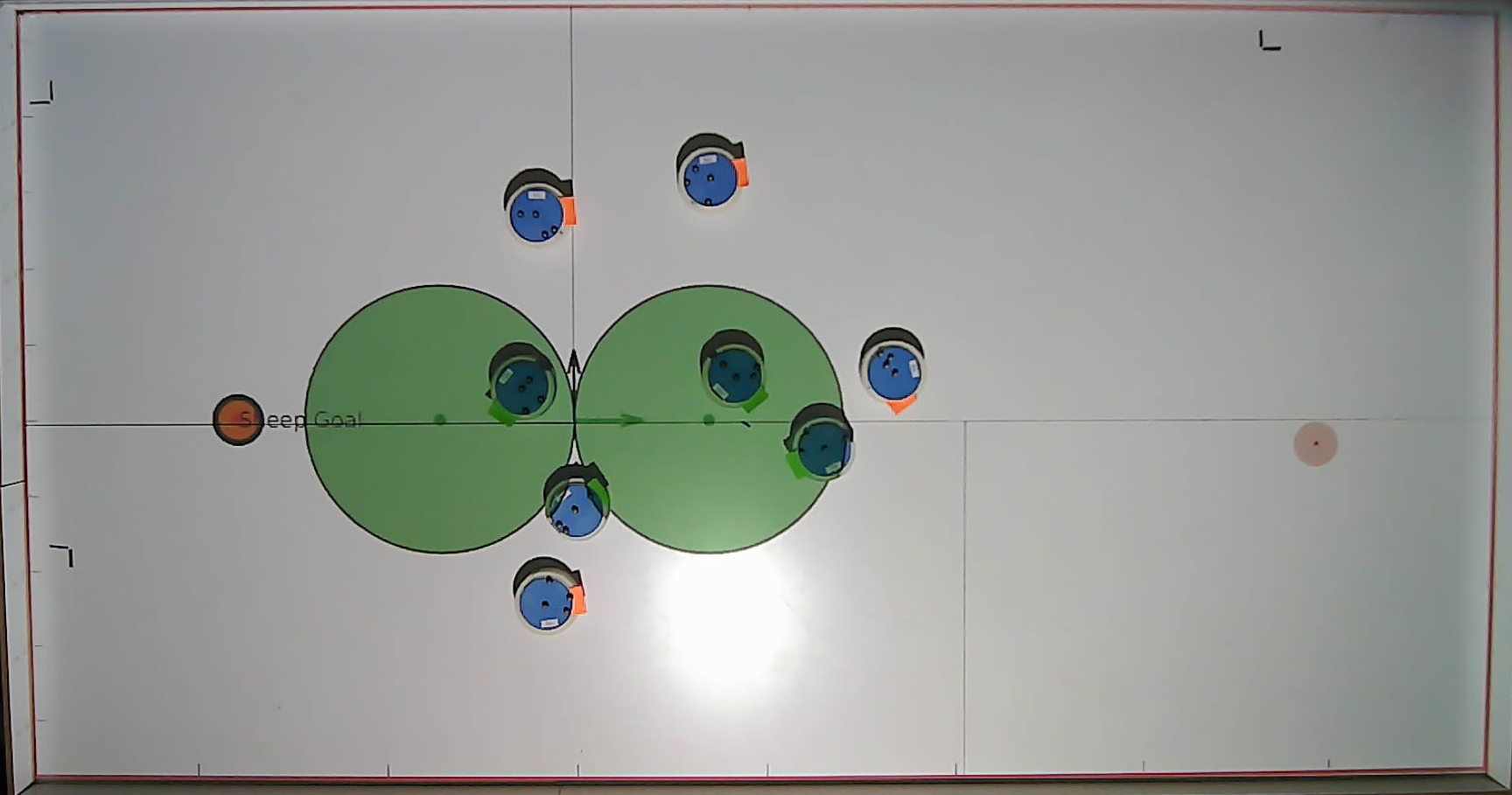}}
	\subfigure[$t = 20s$ ]{\label{fig:dC3}\includegraphics[trim={0.4cm 0.1cm 0.1cm 0cm},clip,width=0.465\columnwidth]{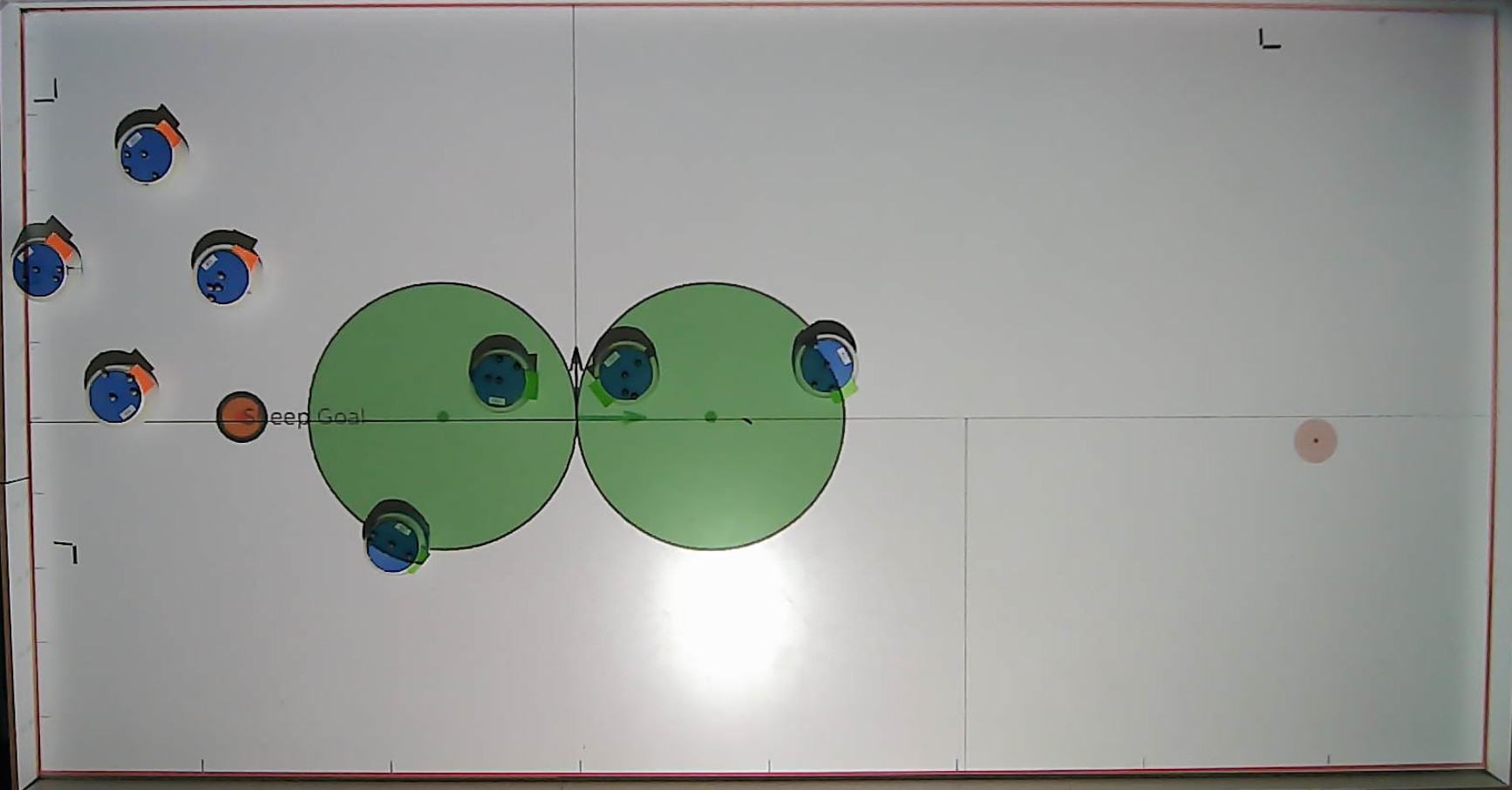}}

	\caption{\textbf{Experiment for the distributed algorithm in section \ref{section:4.1} :} Four dogs (green-tailed robots) defending two protected zone from four sheep (orange-tailed robots).  The goal position $x_G$ (red disc) is in extreme left that would encourage sheep to breach both zones. However, our proposed algorithm moves the dogs so that none of the zones get breached. Video at \url{https://bit.ly/3yo9ziC}.}
	\label{fig:4v4Dcen}

	\vspace{0.4ex}

	\subfigure[$t = 0s$ ]{\label{fig:dA3}\includegraphics[trim={0.4cm 0.1cm 0.1cm 0cm},clip,width=0.465\columnwidth]{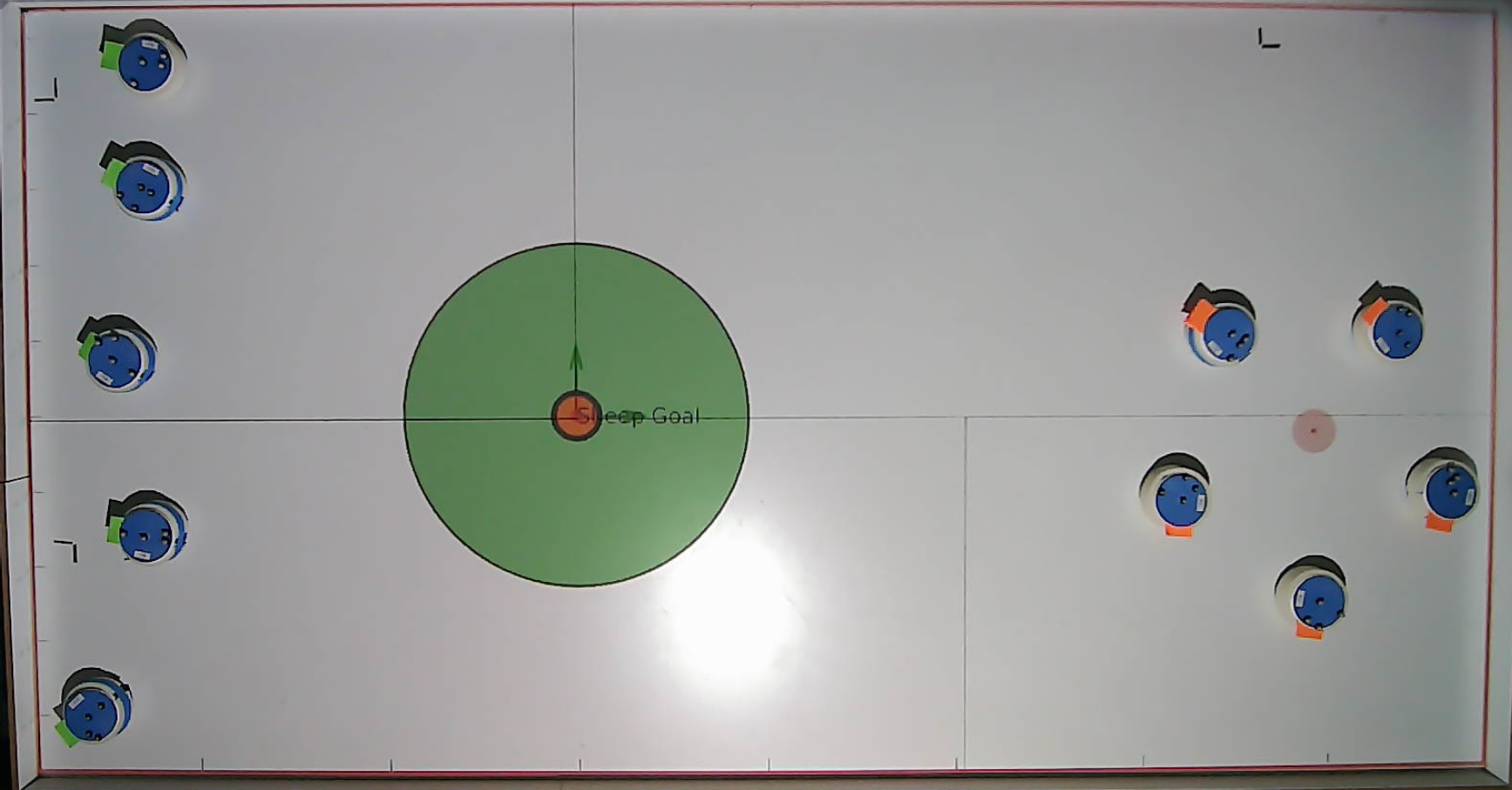}}
	\subfigure[$t = 12s$ ]{\label{fig:dB3}\includegraphics[trim={0.4cm 0.1cm 0.1cm 0cm},clip,width=0.465\columnwidth]{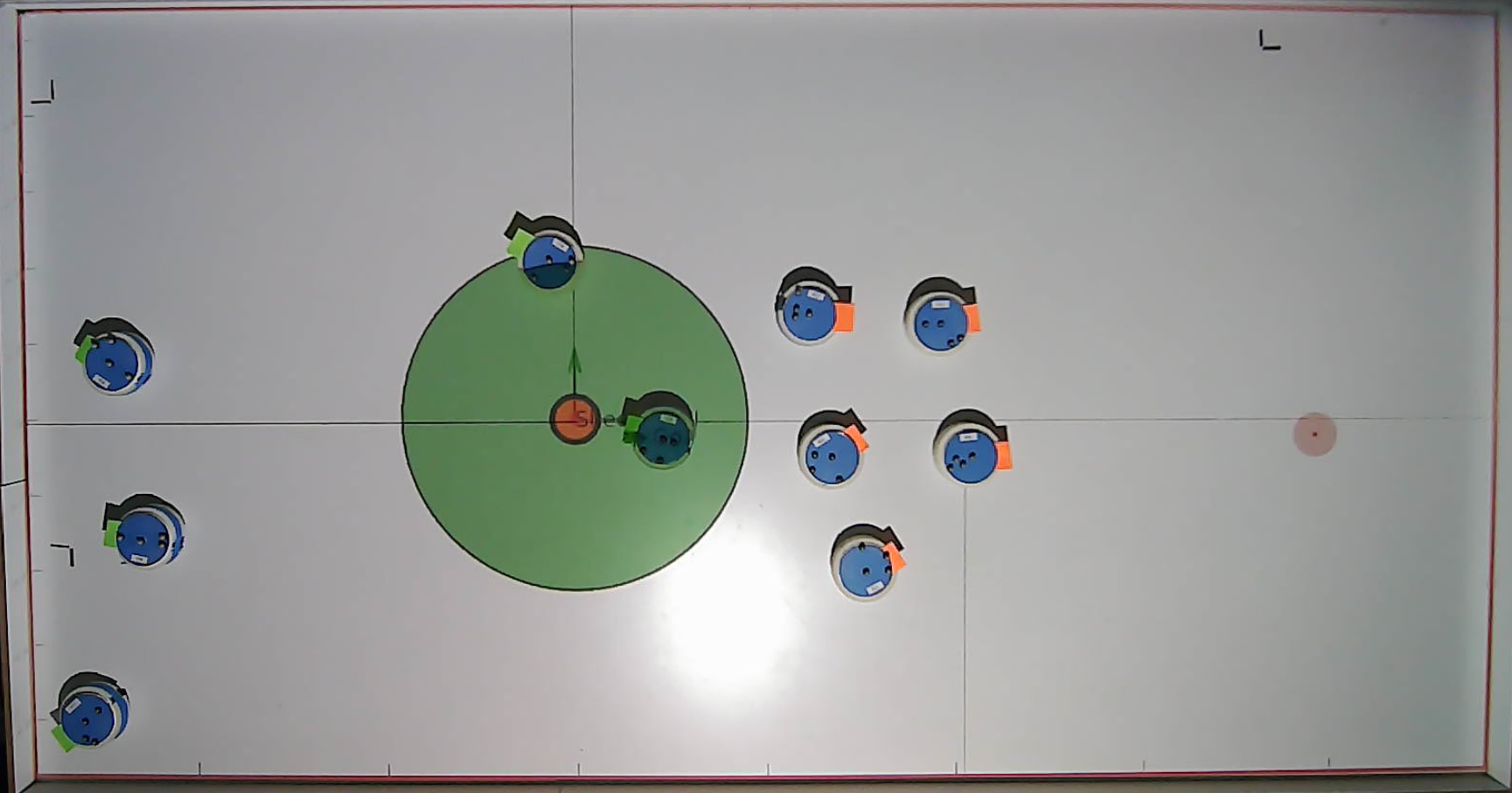}}
	\subfigure[$t = 25s$ ]{\label{fig:dC3}\includegraphics[trim={0.4cm 0.1cm 0.1cm 0cm},clip,width=0.465\columnwidth]{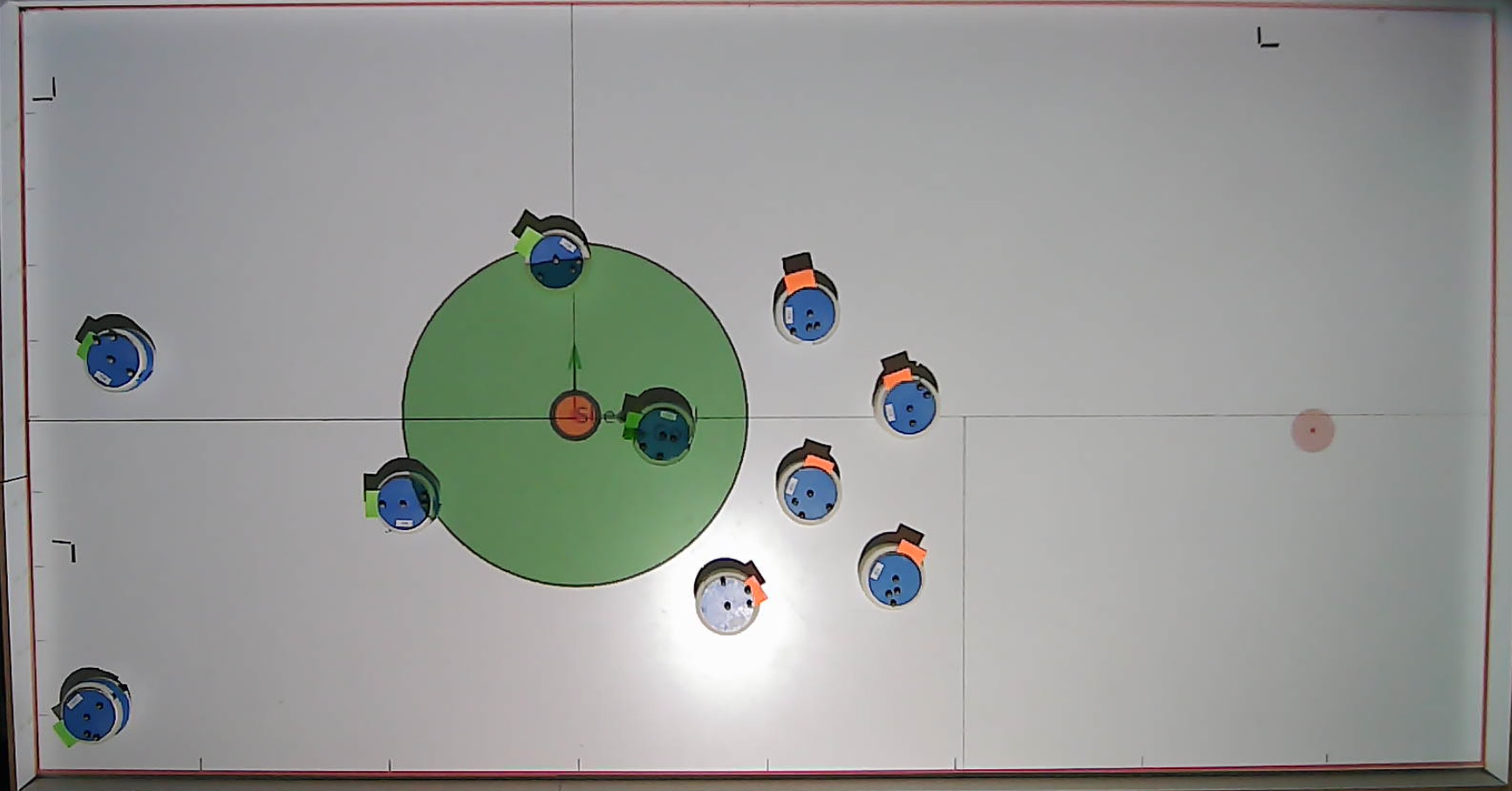}}
	\subfigure[$t = 40s$ ]{\label{fig:dC3}\includegraphics[trim={0.4cm 0.1cm 0.1cm 0cm},clip,width=0.465\columnwidth]{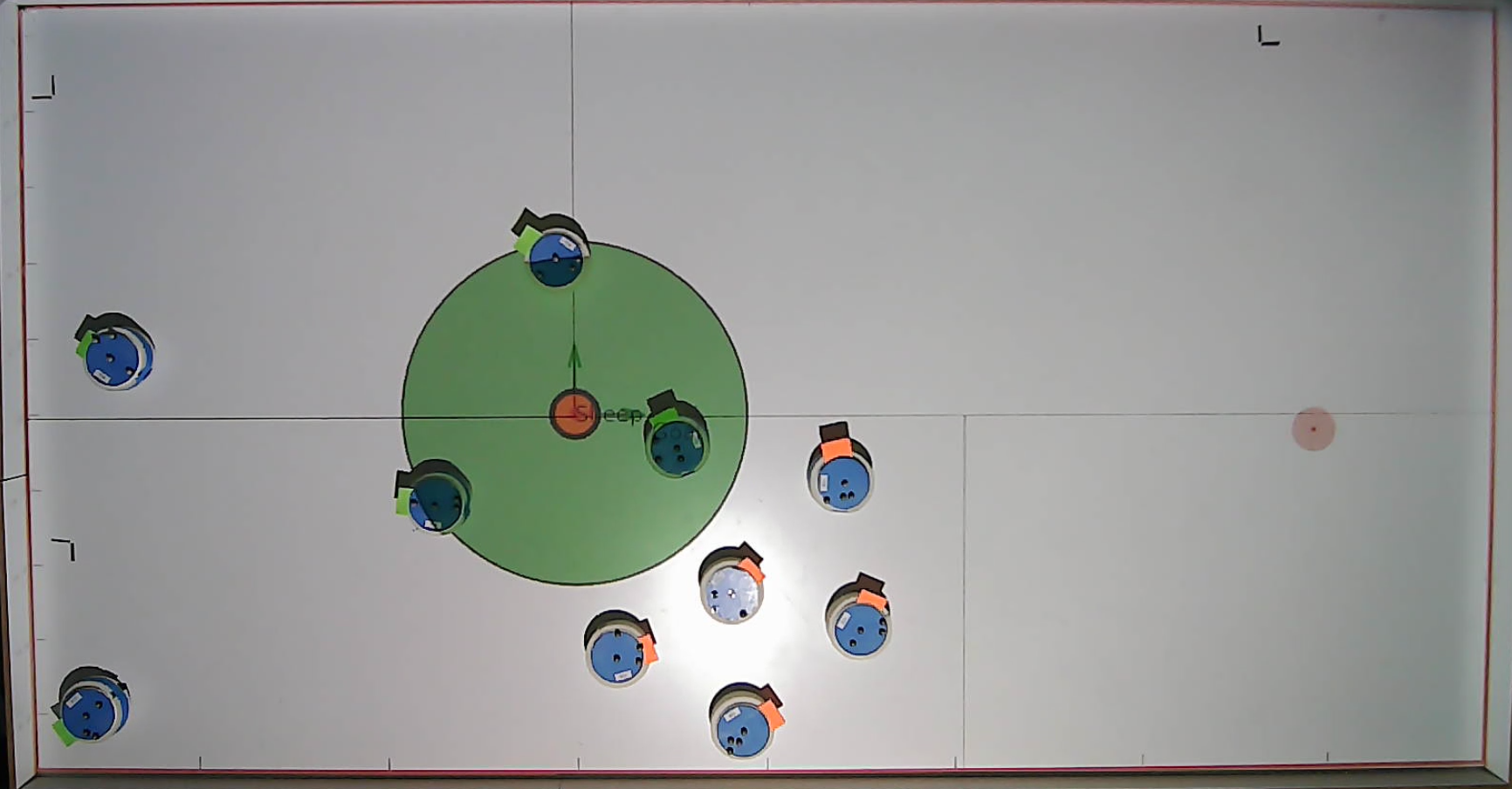}}

	\caption{\textbf{Experiment for the distributed algorithm in section \ref{section:4.1}) :} Five dogs (green-tailed robots) defending the protected zone from five sheep (orange-tailed robots).  The sheep's goal (red disc) is in the center of the protected zone. Eventually, in this scenario a deadlock occurs where all sheep come to a stop outside the protected zone. Video at \url{https://bit.ly/3o51Cu1}. }
	\label{fig:5v5Dcen}
\end{figure*}

\begin{figure*}
    \vspace{0.6ex}

	\centering     
	\subfigure[$t = 0s$ ]{\label{fig:dA3}\includegraphics[trim={0.4cm 0.1cm 0.1cm 0cm} , clip,width=0.465\columnwidth]{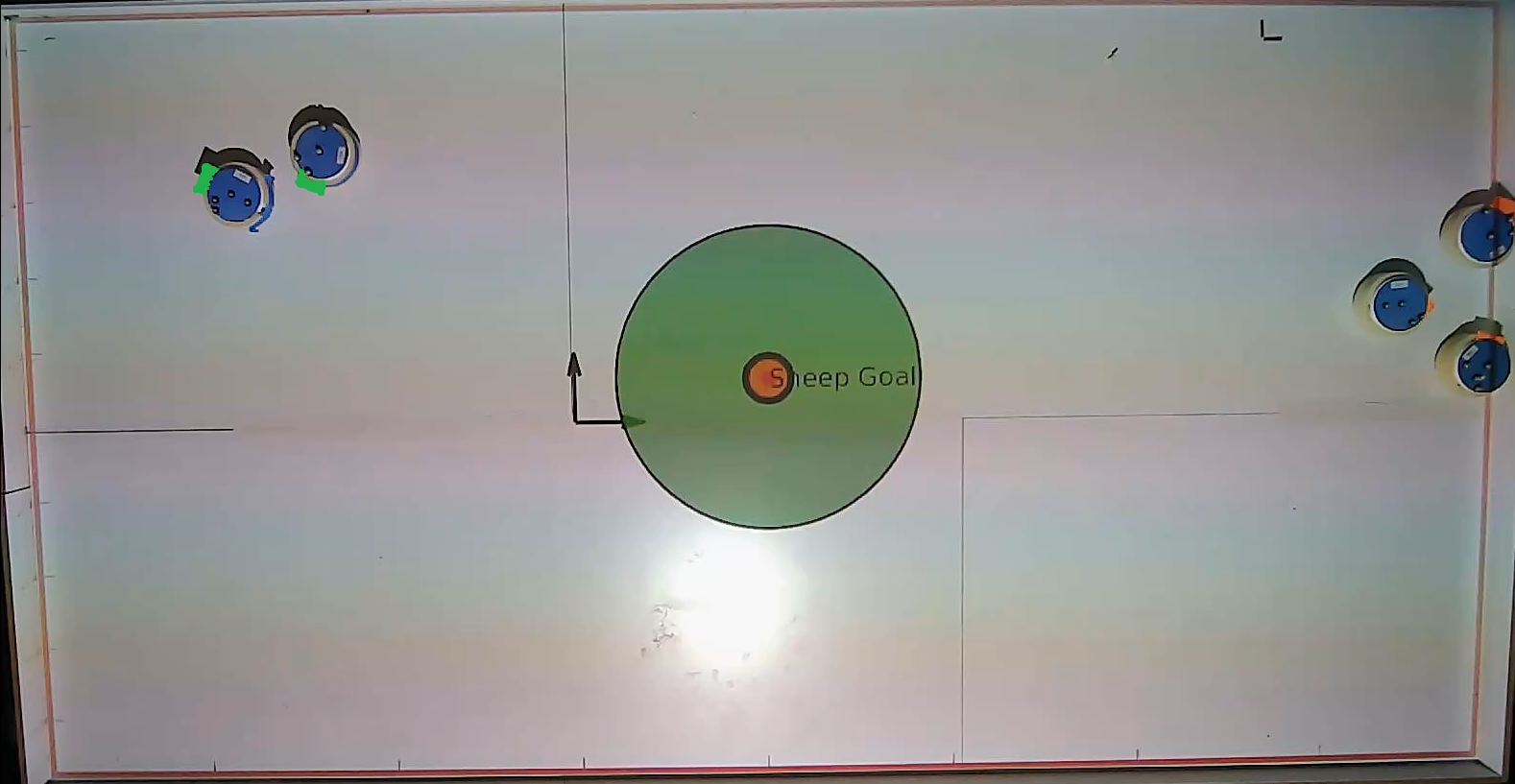}}
	\subfigure[$t = 4s$ ]{\label{fig:dB3}\includegraphics[trim={0.4cm 0.1cm 0.1cm 0cm},clip,width=.465\columnwidth]{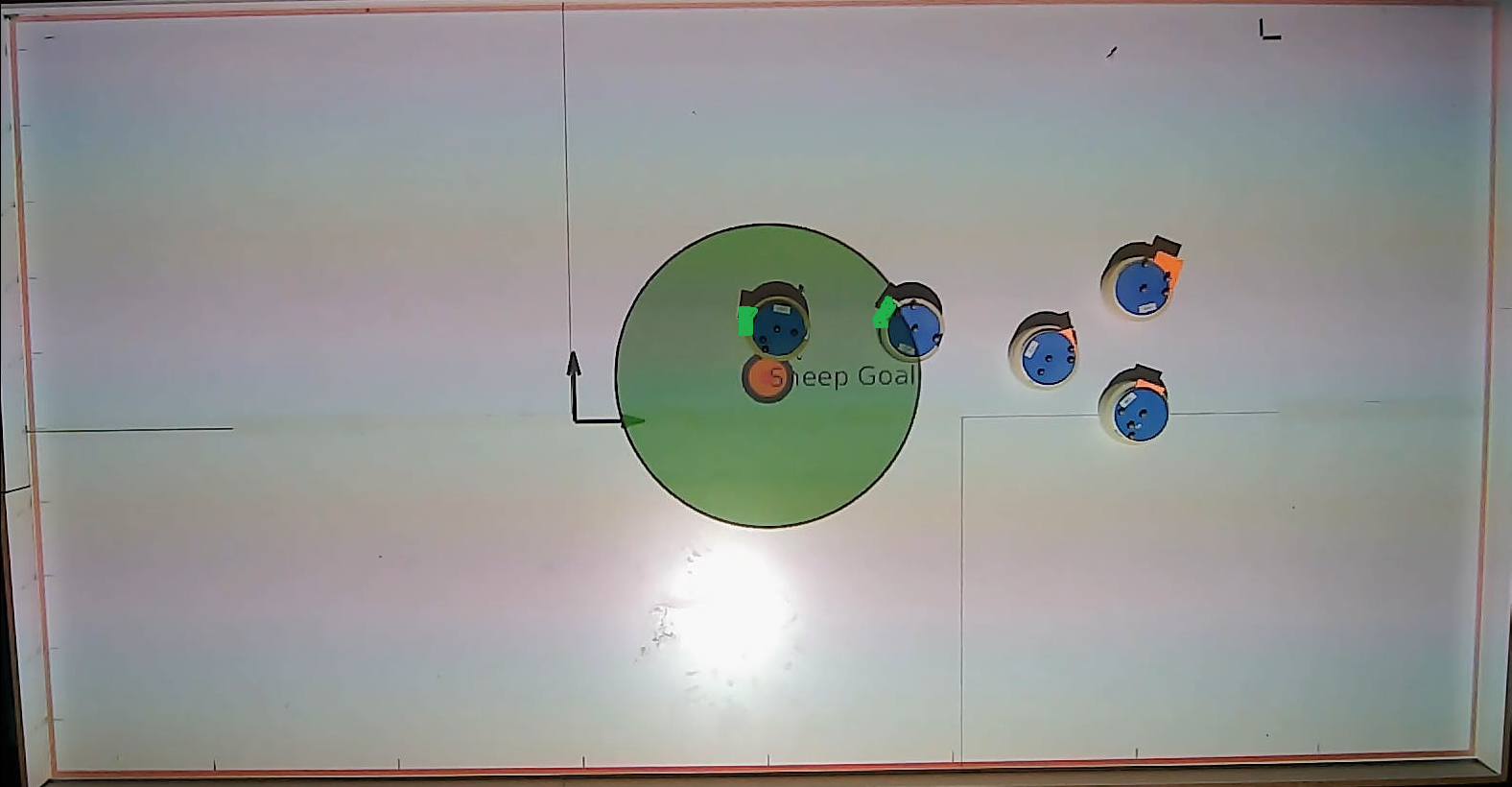}}
	\subfigure[$t = 15s$ ]{\label{fig:dC3}\includegraphics[trim={0.4cm 0.1cm 0.1cm 0cm},clip,width=.465\columnwidth]{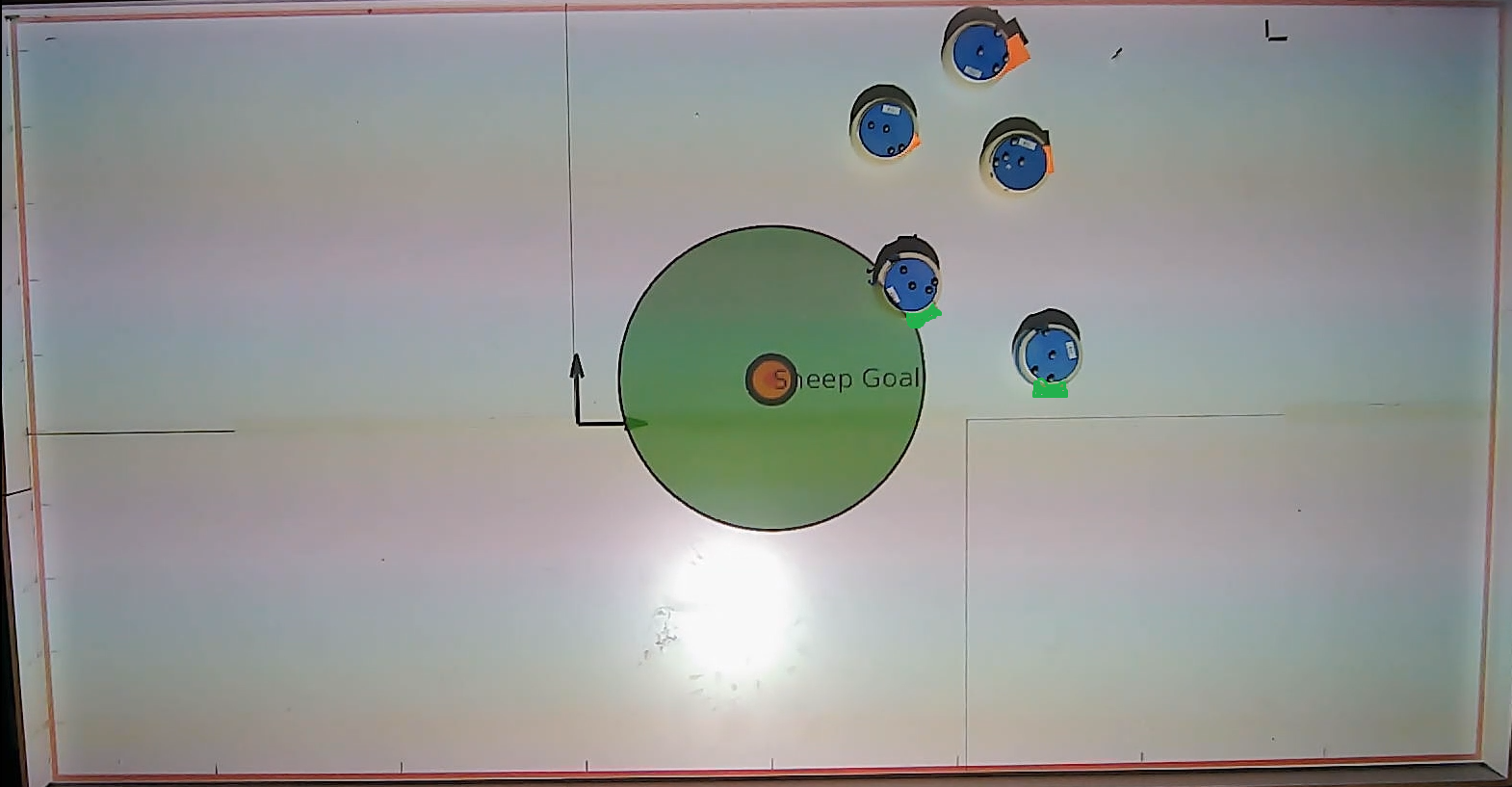}}
	\subfigure[$t = 30s$ ]{\label{fig:dC3}\includegraphics[trim={0.4cm 0.1cm 0.1cm 0cm},clip,width=.465\columnwidth]{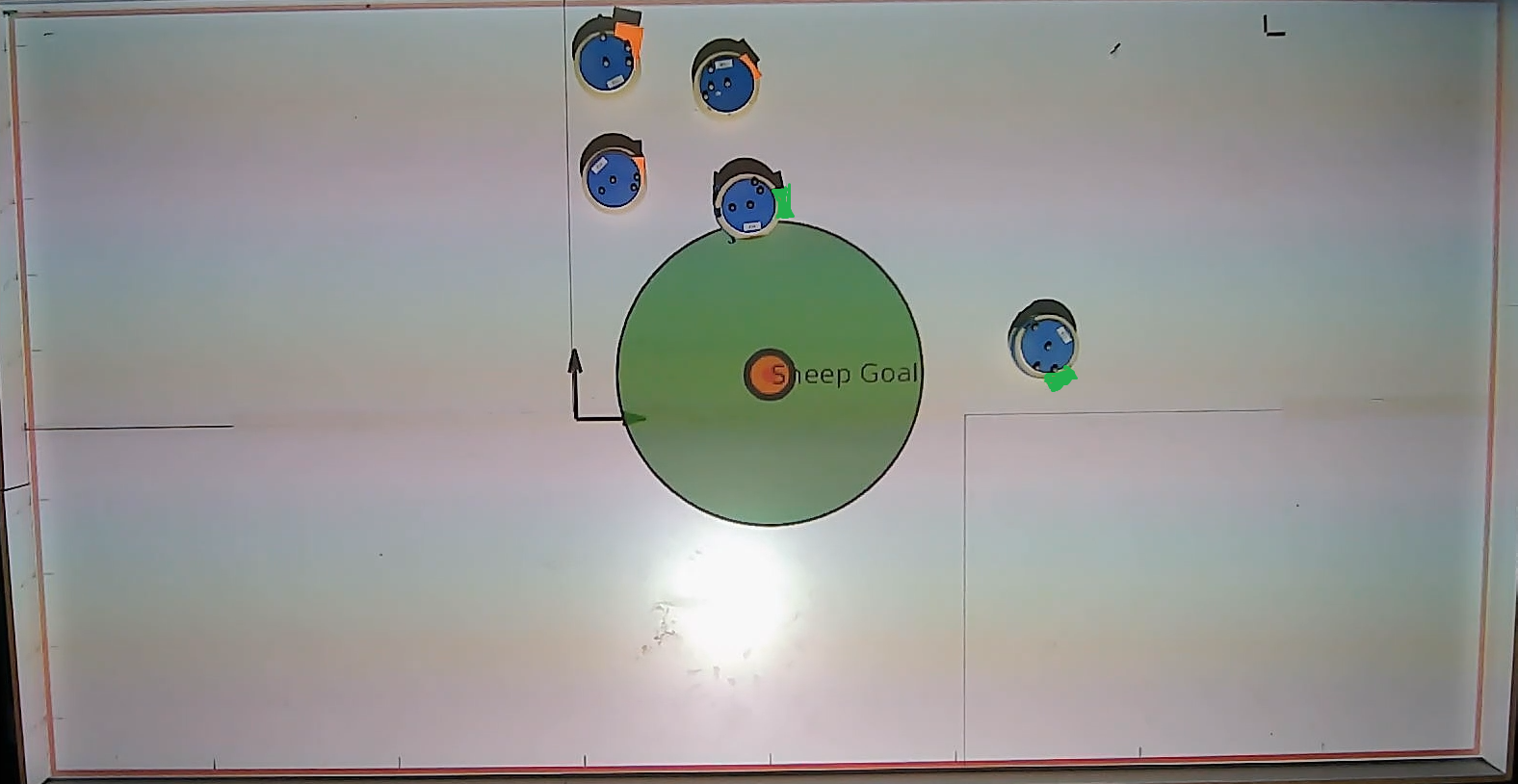}}
	\caption{\textbf{Experiment for distributed algorithm in section \ref{section:4.2}) :} Two dogs (green-tailed robots) defending the protected zone from three sheep (orange-tailed robots).  The goal position $x_G$ (red disc) is at the center of the zone. Video at \url{https://youtu.be/IbCjkR1ye0c}.}
	\label{fig:2v3Dist}


	\centering     
	\subfigure[$t = 0s$ ]{\label{fig:dA3}\includegraphics[trim={0.4cm 0.1cm 0.1cm 0cm} , clip,width=0.465\columnwidth]{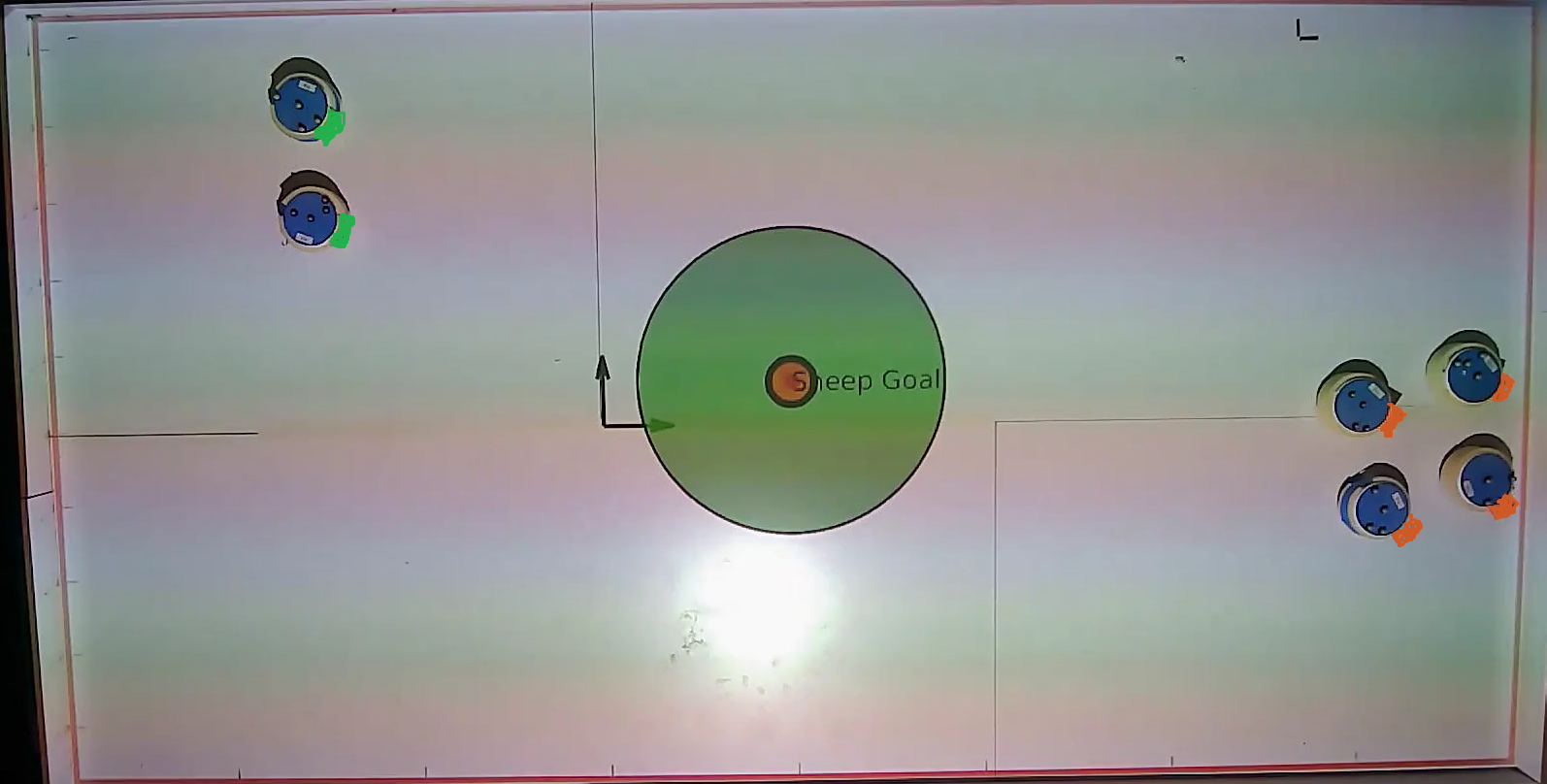}}
	\subfigure[$t = 4s$ ]{\label{fig:dB3}\includegraphics[trim={0.4cm 0.1cm 0.1cm 0cm},clip,width=.465\columnwidth]{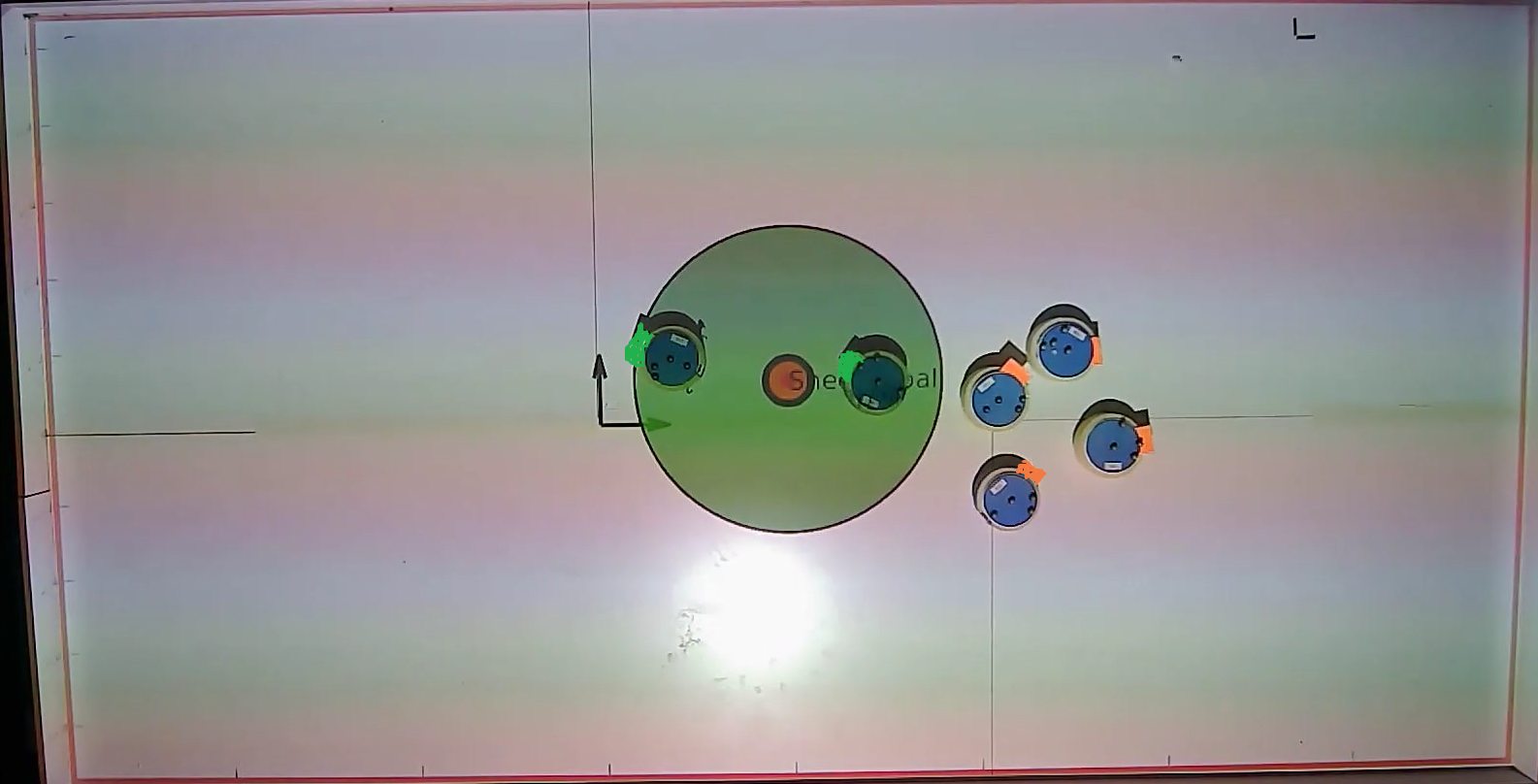}}
	\subfigure[$t = 15s$ ]{\label{fig:dC3}\includegraphics[trim={0.4cm 0.1cm 0.1cm 0cm},clip,width=.465\columnwidth]{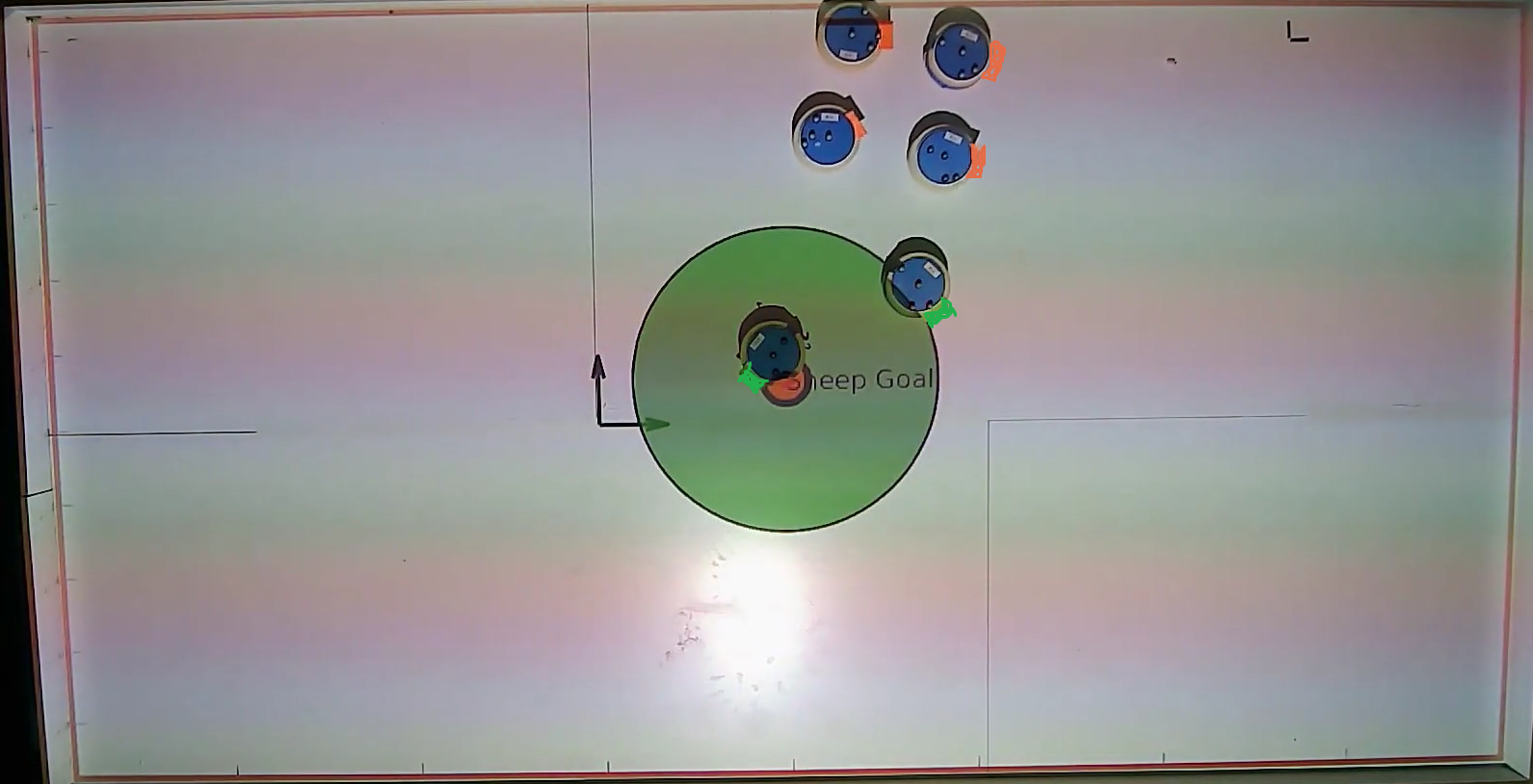}}
	\subfigure[$t = 30s$ ]{\label{fig:dC3}\includegraphics[trim={0.4cm 0.1cm 0.1cm 0cm},clip,width=.465\columnwidth]{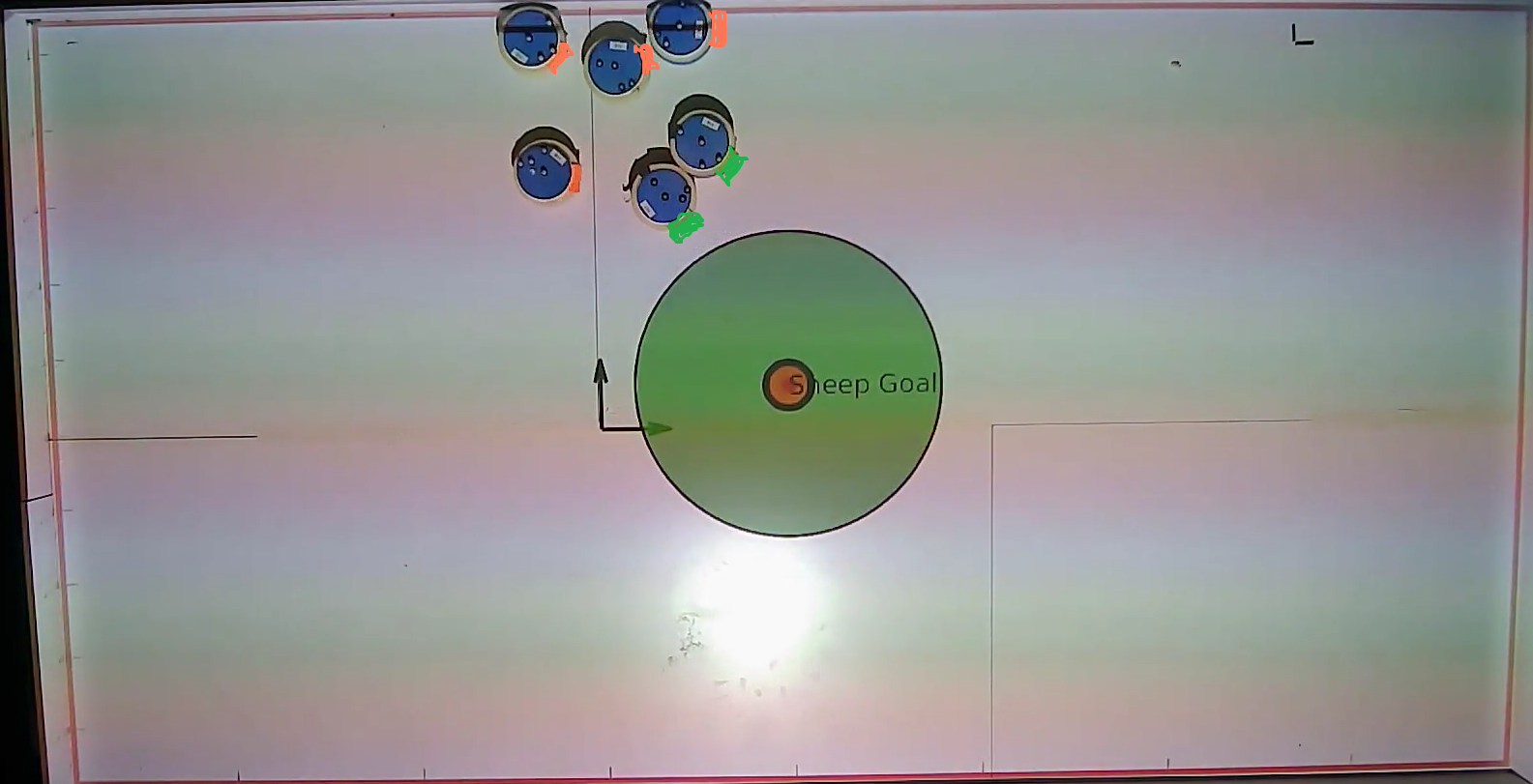}}
	\caption{\textbf{Experiment for distributed algorithm in section \ref{section:4.2}) :} Two dogs (green-tailed robots) defending the protected zone from four sheep (orange-tailed robots). This case is similar to the one shown in fig. \ref{fig:2v4Cen}. Video at \url{https://youtu.be/51FoHZWFYC4}. }
	\label{fig:2v4Dist}
\end{figure*}

Following that, multiple experiments were conducted using the distributed algorithm presented in section \ref{section:4.1}, which requires equal numbers of dogs and sheep. Figure \ref{fig:4v4Dcen} shows 4 dog robots against 4 sheep robots scenario. Here we take two protected zones and show that the dogs can protect both of them. This highlights the compositional nature of our proposed algorithm. We conducted experiments with 5 dog robots and 5 sheep robots, as shown in Figure \ref{fig:5v5Dcen}. Here we can see some dog robots did not require to move as the assigned sheep were being prevented from entering the protected zone due to the configuration of the flock itself. Finally, we test our distributed algorithm presented in section \ref{section:4.2}. Figure \ref{fig:2v3Dist} shows a case where 2 dogs prevent the breaching of protected zone against three dogs. This highlights that our distributed approach can handle under-actuated scenarios. Figure \ref{fig:2v4Dist} and figure \ref{fig:2v4Cen} can be compared to see both centralized and distributed algorithm handling a similar scenario of 2 dogs against 4 sheep.

\section{Conclusions}
\label{conclusions}
\vspace{-0.75ex}
In this paper, we developed a novel optimization-based distributed control techniques to enable multiple dog robots to prevent the breaching of protected zones by sheep agents. We provided proof of feasibility of the controller when $n$ dog robots face an equal number of sheep robots. Additionally, we developed another distributed algorithm that iteratively computes a solution that agrees with the solution returned by the centralized problem without requiring equal number of dogs and sheep. We experimentally validated both distributed algorithms in addition to validating our previously developed centralized control. We show that multiple dog robots can prevent breaching of protected zone in both simulation and real-world experiments. In future work, we aim for the dog robots to learn the dynamics of the sheep robots online while preventing them from breaching. 
\newpage
\section{Appendix: Proof of feasibility for Approach 1}

\vspace{-0.5ex}

\label{dist_appendix}
\setcounter{equation}{0}
\setcounter{theorem}{0}
\begin{theorem}
\label{theorem_caseB} In a scenario with `$n$' dogs and `$n$' sheep, with each dog assigned a unique sheep, the herding constraint \eqref{herdingcon1} for a given dog is always feasible, provided assumptions \ref{ass4} are met.
\end{theorem}

\vspace{-2.5ex}

\begin{proof}
Our strategy to guarantee feasibility of constraint \eqref{herdingcon1} relies on ruling out situations in which it is infeasible. \eqref{herdingcon1} can become infeasible

\vspace{-0.5ex}

\begin{itemize}
    \item  either when $A^H_i = \boldsymbol{0}$ and $b^H_i<0$ (\textcolor{blue}{possibility 1})
    \item or when $b^H_i = -\infty$ (\textcolor{blue}{possibility 2}).
\end{itemize} 

\vspace{-0.5ex}

\noindent To determine the conditions in which \textcolor{blue}{possibility 1} occurs, we calculate the determinant of $\mathbb{J}_{ki}^D$ as

\vspace{-5ex}

\begin{align*}
    det(\mathbb{J}_{ki}^D) = \frac{-2k_D^2}{\| \boldsymbol{x}_{D_k} -\boldsymbol{x}_{S_i}\|^3} 
\end{align*}
The determinant $det(\mathbb{J}_{ki}^D)$ is non-zero as long as the distance between dog $D_k$ and sheep $S_i$ is finite. Therefore,  $\mathbb{J}_{ki}^D$ will have no null space,  implying that $A^H_i \neq 0$ $\forall \boldsymbol{x}_{S_i} \in \mathbb{R}^2, \boldsymbol{x}_{D_k} \in \mathbb{R}^2$.
This rules out \textcolor{blue}{possibility 1} for infeasibility. To rule out \textcolor{blue}{possibility 2}, we need to check for condition when  $b^H_i \longrightarrow -\infty $. Given $b^H_i$ in \eqref{herdingcon1}, we find its worst case lower bound. Here $ \boldsymbol{f}_i^T\boldsymbol{f}_i \geq 0$ and as we assume that at the current time step, the sheep is outside $\mathcal{P}$, this ensures $\beta \frac{h}{2} \geq 0$. By removing these terms, the lower bound of $b^H_i$ can be given as
\begin{align}
\label{eq:ineq}
b^H_i  \geq& \sum_{j \in \mathcal{S}\backslash i} \left(\boldsymbol{x}_{S_i} - \boldsymbol{x}_P\right)^T\mathbb{J}_{ji}^S \boldsymbol{f}_j 
+ (\boldsymbol{x}_{S_i} - \boldsymbol{x}_P)^T\mathbb{J}_{ii}^S \boldsymbol{f}_i
+ \sum_{l \in \mathcal{D}\backslash k} (\boldsymbol{x}_{S_i} - \boldsymbol{x}_P)^T\mathbb{J}_{li}^D \boldsymbol{u}_{D_l} \nonumber
\\&+ \alpha (\boldsymbol{x}_{S_i} - \boldsymbol{x}_P)^T  \boldsymbol{f}_i
\end{align}
Using the triangle inequality on the RHS and Cauchy-Schwarz inequality on individual terms, we get
\begin{align}
b^H_i \geq & \sum_{j \in \mathcal{S}\backslash i} \left(-\sigma_{max}\left(\mathbb{J}_{ji}^S\right) \left\|\boldsymbol{x}_{S_{i}}-\boldsymbol{x}_P\right\| \|\boldsymbol{f}_{j} \| \right)
-\sigma_{max}\left(\mathbb{J}_{ii}^S\right) \left\|\boldsymbol{x}_{S_{i}}-\boldsymbol{x}_P\right\| \|\boldsymbol{f}_{i} \| \\
& + \sum_{l \in \mathcal{D}\backslash k} \left( -\sigma_{max}\left(\mathbb{J}_{li}^D\right) \left\|\boldsymbol{x}_{S_{i}}-\boldsymbol{x}_P\right\|\left\|\boldsymbol{u}_{D_{l}}\right\| \right)
-\alpha\| \boldsymbol{x}_{S_{i}}-\boldsymbol{x}_P\|\| \boldsymbol{f}_{i} \| \notag
\end{align}
where $\sigma_{max}$ is the largest singular value of a matrix.  
Further, using the fact that the largest singular value of a matrix ($\sigma_{max}$) is upper bounded by its Frobenius norm ($\sigma_F$), we obtain
\begin{align}
    \label{eq:bh1ineq4}
b^H_i \geq &\sum_{j \in \mathcal{S}\backslash i} \left(-\sigma_{F}\left(\mathbb{J}_{ji}^S\right) \left\|\boldsymbol{x}_{S_{i}}-\boldsymbol{x}_P\right\| \|\boldsymbol{f}_{j} \| \right)
-\sigma_{F}\left(\mathbb{J}_{ii}^S\right) \left\|\boldsymbol{x}_{S_{i}}-\boldsymbol{x}_P\right\| \|\boldsymbol{f}_{i} \| \\
& \sum_{l \in \mathcal{D}\backslash k} \left( -\sigma_{F}\left(\mathbb{J}_{li}^D\right) \left\|\boldsymbol{x}_{S_{i}}-\boldsymbol{x}_P\right\|\left\|\boldsymbol{u}_{D_{l}}\right\| \right)
-\alpha\| \boldsymbol{x}_{S_{i}}-\boldsymbol{x}_P\|\| \boldsymbol{f}_{i} \| \notag
\end{align}
Now to compute this lower bound we make use of assumption \ref{ass4}. We use the dynamics in  \eqref{sheepdynamics} to compute $\mathbb{J}_{ii}^S$ and obtain the upper bound on $\sigma_{F}\left(\mathbb{J}_{ii}^S\right)$
and use the bounds on distances from assumption \ref{ass4} to get
following upper bound:

\vspace{-4ex}

\begin{align*}
\label{frobnorm}
\sigma_{F}\left(\mathbb{J}_{ii}^S\right) &\leqslant  \sum_{j \in \mathcal{S}\backslash i} k_S\left( \sqrt{2} +\frac{(3+\sqrt{2}) R^{3}}{\| \boldsymbol{x}_{S_{i}}-\boldsymbol{x}_{S_{j}}\|^{3}} \right) + \sqrt{2}k_G  
+\sum_{l \in \mathcal{D}} \frac{\left(3+\sqrt{2}\right) k_{D}}{\left\|\boldsymbol{x}_{S_{i}}-\boldsymbol{x}_{D_{l}}\right\|^{3}}\\
&\leqslant  (n-1)k_S\left( \sqrt{2} +\frac{(3+\sqrt{2}) R^{3}}{L_S^{3}} \right) + \sqrt{2}k_G  
+ n \left(\frac{\left(3+\sqrt{2}\right) k_{D}}{L_D^{3}} \right) \coloneqq \lambda_M
\end{align*}

\vspace{-1.2ex}

\noindent We omit the proof of this computation in the interest of space.
Similarly, using the dynamics  in \eqref{sheepdynamics}, we compute an expression for $\mathbb{J}_{ji}^S$ and obtain an upper bound on  $\sigma_{F}\left(\mathbb{J}_{ji}^S\right)$ as follows:

\vspace{-4ex}

\begin{align*}
\sigma_{F}\left(\mathbb{J}_{ji}^S\right)  \leqslant   \sqrt{2} k_{S}+\frac{(3+\sqrt{2}) k_{S} R^{3}}{\| \boldsymbol{x}_{S_{i}}-\boldsymbol{x}_{S_{j}}\|^{3}}  \leqslant \sqrt{2} k_{S}+\frac{(3+\sqrt{2}) k_{S} R^{3}}{L_S^{3}} \coloneqq \lambda_S
\end{align*}
\noindent Likewise, an upper bound of $\sigma_{F}\left(\mathbb{J}_{li}^D\right)$, is given by
\begin{align*}
\sigma_{F}\left(\mathbb{J}_{li}^D\right)  \leqslant   \frac{(3+\sqrt{2}) k_{D}}{\| \boldsymbol{x}_{S_{i}}-\boldsymbol{x}_{D_{l}}\|^{3}}  &\leqslant \frac{(3+\sqrt{2}) k_{D}}{L_D^{3}} \coloneqq \lambda_D 
\end{align*}
Lastly, we use obtain an upper bound on the dynamics of each sheep $\boldsymbol{f}_i$ as:
\begin{align}
    \|\boldsymbol{f}_i\| \leqslant& 
    \sum_{j \in \mathcal{S}\backslash i} k_S\left(\|\boldsymbol{x}_{S_i}-\boldsymbol{x}_{S_j}\| + \frac{R^3}{\|\boldsymbol{x}_{S_i}-\boldsymbol{x}_{S_j}\|^2}\right)
    +k_{G}\|\boldsymbol{x}_{G}-\boldsymbol{x}_{S_i}\| \nonumber \\
    &+ \sum_{l \in \mathcal{D}} k_{D}  \frac{\|\boldsymbol{x}_{S_i} - \boldsymbol{x}_{D_l}\|}{\|\boldsymbol{x}_{S_i} - \boldsymbol{x}_{D_l}\|^{3}}
\end{align}
Now we need to compute the maximum possible value of the RHS to get the upper bound of the sheep dynamics. The first term has a local minima at $\|\boldsymbol{x}_{S_i} - \boldsymbol{x}_{S_j}\| = (2)^{1/3}R$. Therefore the maximum value can occur at either the lower bound or upper bound of $\|\boldsymbol{x}_{S_i} - \boldsymbol{x}_{S_j}\|$. Thus the maximum value of the first term can be given as $F_{max} \coloneqq \max (k_SL_S + k_S\frac{R^3}{L_S^2},k_SM_S + k_S\frac{R^3}{M_S^2})$.
Second term is maximum when $\|\boldsymbol{x}_{G} - \boldsymbol{x}_{S_i}\| = M_G$. The last term is maximum when distance of the sheep to the dogs are minimum,  $\|\boldsymbol{x}_{S_i} - \boldsymbol{x}_{D_k}\| = L_D$. Using these the upper bound on the sheep dynamics is computed as:

\vspace{-4ex}

\begin{align*}
    \|\boldsymbol{f}_i\| \leqslant (n-1)F_{max} 
    +k_{G} M_{G}+nk_{D}\left(\frac{1}{L_D^{2}}\right)
\end{align*}

\noindent Assuming that the velocity of the dog robots have an upper bound, and by taking the upper bound on the dynamics of all the sheep to be equal, the lower bound on $b^H_i$ from  \ref{eq:bh1ineq4} is (taking $\gamma = -(\alpha+\lambda_{M}+ (n-1)\lambda_{S})M_p$)
\begin{align*}
    \begin{aligned}
b^{H}_i &\geqslant \gamma \left\{(n-1)F_{max}
    +k_{G} M_{G}+\frac{nk_{D}}{L_D^{2}}\right\}
-(n-1)\lambda_D M_{P}\left\|\boldsymbol{u}_{D}\right\|_{\text {max }}
\end{aligned}
\end{align*}
This shows that $b^H_i$ has a finite lower bound, thus ruling out \textcolor{blue}{possibility 2}. Thus, the herding constraint \eqref{herdingcon1} for a one dog to repel one sheep from the protected zone is always feasible. Since each sheep in $\mathcal{S}$ is allocated to one unique dog in $\mathcal{D}$, extension of this feasibility result to all sheep ensures that none of them will breach the protected zone.
\end{proof}

%
%
\bibliographystyle{IEEEtran}
\bibliography{root}









\end{document}